\documentclass[pdflatex,sn-mathphys-num,iicol]{sn-jnl}

\usepackage{graphicx}%
\usepackage{multirow}%
\usepackage{amsmath,amssymb,amsfonts}%
\usepackage{amsthm}%
\usepackage{mathrsfs}%
\usepackage[title]{appendix}%
\usepackage{xcolor}%
\usepackage{textcomp}%
\usepackage{manyfoot}%
\usepackage{booktabs}%
\usepackage{algorithm}%
\usepackage{algorithmicx}%
\usepackage{algpseudocode}%
\usepackage{listings}%
\usepackage{graphicx} 
\usepackage[table]{xcolor}
\usepackage{textcomp}
\usepackage{url}
\usepackage{hyperref}

\theoremstyle{thmstyleone}%
\theoremstyle{thmstyletwo}%

\theoremstyle{thmstylethree}%

\begin{document}

\title[Article Title]{\textbf{GTA}: Advancing Image-to-3D World Generation via \textbf{G}eometry \textbf{T}hen \textbf{A}ppearance Video Diffusion}


\author[1]{\fnm{Hanxin} \sur{Zhu}}\email{hanxinzhu@mail.ustc.edu.cn}

\author[2,3]{\fnm{Cong} \sur{Wang}}

\author[2,5]{\fnm{Peiyan} \sur{Tu}}

\author[2,6]{\fnm{Jiayi} \sur{Luo}} 
 
\author{\fnm{Tianyu} \sur{He}}
\email{deeptimhe@gmail.com}

\author[2,4]{\fnm{Xin} \sur{Jin}}

\author*[1,2]{\fnm{Zhibo} \sur{Chen}}\email{chenzhibo@ustc.edu.cn}

\affil[1]{\orgdiv{School of Information Science and Technology}, \orgname{University of Science and Technology of China}, \orgaddress{\state{Hefei}, \country{China}}}

\affil[2]{\orgdiv{Zhongguancun Academy}, \orgaddress{\city{Beijing}, \country{China}}}

\affil[3]{\orgdiv{the State Key Laboratory of Multimodal Artificial Intelligence Systems}, \orgname{Institute of Automation, Chinese Academy of Sciences}, \orgaddress{\city{Beijing}, \country{China}}}

\affil[4]{\orgdiv{Eastern Institute of Technology}, \orgaddress{\city{Ningbo}, \country{China}}}

\affil[5]{\orgdiv{College of Information Science and Electronic Engineering, Zhejiang University, Hangzhou, China}}

\affil[6]{\orgdiv{School of Computer Science and Engineering}, \orgname{Beihang University}, \orgaddress{\city{Beijing}, \country{China}}}





\abstract{
Recent developments in generative models and large-scale datasets have substantially advanced 3D world generation, facilitating a broad range of domains including spatial intelligence, embodied intelligence, and autonomous driving. While achieving remarkable progress, existing approaches to 3D world generation typically prioritize appearance prediction with limited modeling of the underlying geometry, leading to issues such as unreliable scene structure estimation and degraded cross-view consistency. To address these limitations, motivated by the coarse-to-fine nature of human visual perception, we propose \textbf{GTA}, a novel image-to-3D world generation method following a \textbf{G}eometry-\textbf{T}hen-\textbf{A}ppearance paradigm. Specifically, given a single input image, to improve the structural fidelity of synthesized 3D scenes, GTA adopts a two-stage framework with two dedicated video diffusion models, which first generate coarse geometric structure from novel viewpoints and then synthesize fine-grained appearance conditioned on the predicted geometry. To further enhance cross-view appearance consistency, we introduce a random latent shuffle strategy during the training process, along with a test-time scaling scheme that improves perceptual quality without compromising quantitative performance. Extensive experiments have demonstrated that our proposed method consistently outperforms existing approaches in terms of fidelity, visual quality, and geometric accuracy. Moreover, GTA is shown to be effective as a general enhancement module that further improves the generation quality of existing image-to-3D world pipelines, as well as supporting multiple downstream applications and exhibiting favorable data efficiency during model training, highlighting its versatility and broad applicability.
Project page: \url{https://hanxinzhu-lab.github.io/GTA/}.
}


\begin{figure*}[t]
\centering
\includegraphics[width=0.9\textwidth]{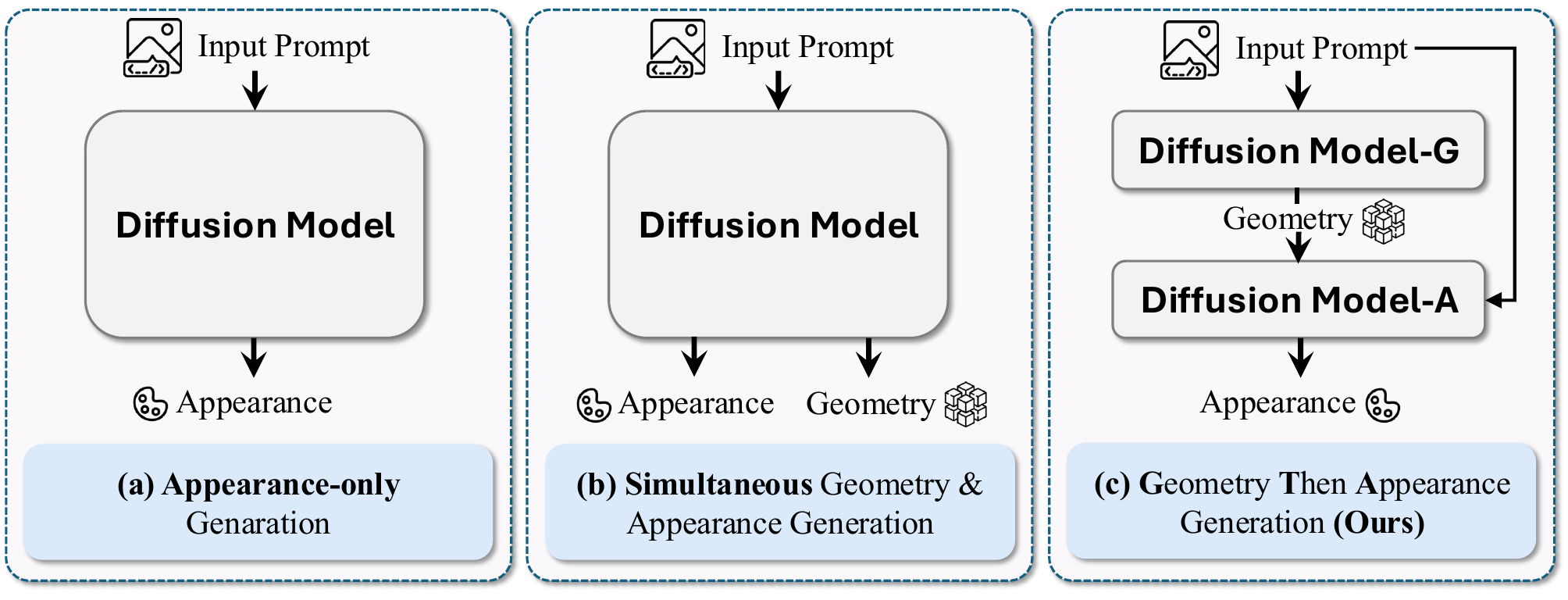}
\caption{\textbf{Comparisons of pipelines for 3D world generation.} \textbf{(a)} Appearance-only generation, which focuses on synthesizing appearance without modeling scene geometry.\textbf{(b)} Simultaneous geometry and appearance generation, which jointly predicts both modalities within a single diffusion model.
\textbf{(c)} \textbf{G}eometry-\textbf{T}hen-\textbf{A}ppearance generation \textbf{(ours)}, which first estimates multi-view geometric structure and subsequently synthesizes appearance conditioned on the predicted geometry.}
\label{fig1}
\end{figure*}



\maketitle

\section{Introduction}\label{sec1}
As the fundamental medium for human perception, understanding, and action, 3D spatial environments underpin all activities in the physical world. For decades, the digital representation, reconstruction, and generation of such complex environments have remained a central and long-standing challenge in computer vision, computer graphics, and artificial intelligence~\cite{butime20063d,li2024advances,peng2023openscene,samavati2023deep,wen20253d,zia2015towards}. As one of the key research directions, 3D generation~\cite{li2024advances,wen20253d, li2024instant3d,moschoglou20203dfacegan,di2025hyper,song2023automatic} focuses on synthesizing 3D assets with faithful spatial structure and appearance from multimodal inputs such as text and images. By bridging the physical and digital worlds, it serves as a foundation for a broad spectrum of domains, ranging from embodied intelligence~\cite{yang2024holodeck,wang2025embodiedgen} to world modeling~\cite{kong20253d,ding2025understanding,zheng2024occworld}.

Early approaches to 3D generation primarily focused on object-level synthesis~\cite{poole2022dreamfusion,lin2023magic3d,raj2023dreambooth3d,wang2023score}, where individual 3D objects or simple scenes were generated by distilling priors from pretrained generative models. While effective for isolated objects, such object-centric formulations are inherently limited in scalability and struggle to capture the complex spatial layouts and long-range relationships present in real-world scenes. Recently, driven by the powerful spatiotemporal modeling capabilities of video generative models~\cite{hong2022cogvideo,gupta2024photorealistic,bar2024lumiere}, together with the availability of large-scale internet data~\cite{nan2024openvid, wang2023internvid}, research has increasingly shifted towards scene-level 3D generation~\cite{sun2024dimensionx,bahmani2025ac3d,liang2025wonderland}, aiming to synthesize complex 3D worlds with coherent geometry and consistent appearance across different viewpoints.

Despite achieving encouraging progress, as shown in Fig.~\ref{fig1}(a), most existing scene-level approaches remain largely appearance-centric~\cite{yu2024viewcrafter,zhang2025spatialcrafter,he2024cameractrl,wang2024motionctrl}, exhibiting a pronounced bias towards texture synthesis with limited modeling of the underlying geometry. However, since geometric structure provides the fundamental spatial scaffold that constrains scene layout and enforces cross-view consistency, insufficient geometric modeling leaves appearance synthesis inherently under-constrained, typically resulting in unreliable scene structure estimation and view-inconsistent renderings. These observations suggest that, for 3D world generation, the construction of geometry and appearance is equally essential, and both should be carefully considered to achieve a faithful and consistent 3D scene.

To this end, a straightforward way is to simultaneously predict geometry and appearance within a unified framework~\cite{huang2025voyager,dai2025fantasyworld}, aiming to obtain a coherent scene representation, as shown in Fig.~\ref{fig1}(b). However, simultaneous prediction is inherently challenging~\cite{zhang2025dualcamctrl}, as geometry and appearance differ substantially in learning difficulty, data distributions, and uncertainty. As a result, optimizing both modalities at the same time often induces mutual interference, leading to geometric distortions, appearance flickering, and degraded cross-view consistency.

In nature, human visual perception has been extensively characterized as exhibiting a coarse-to-fine organization~\cite{navon1977forest,watt1987scanning}, wherein global scene structure is processed prior to fine-scale surface details. Inspired by this perceptual principle, we posit that 3D world generation should similarly prioritize the establishment of geometric structure before appearance modeling, thereby reducing ambiguity and alleviating the interference arising from simultaneous prediction. 

\textcolor{black}{Guided by this perspective, in this paper we propose \textbf{GTA}, a novel image-to-3D world generation framework that follows a \textbf{G}eometry-\textbf{T}hen-\textbf{A}ppearance paradigm, where geometry is modeled as a learned generative intermediate to condition subsequent appearance synthesis.} As demonstrated in Fig.~\ref{fig1}(c), leveraging two dedicated video diffusion models (referred to as Diffusion Model-G and Diffusion Model-A), the framework first generates a coarse yet globally consistent geometric representation of the scene across novel viewpoints. This geometry-aware representation then serves as an explicit structural prior to guide a subsequent appearance synthesis stage, enabling fine-grained texture generation under well-defined spatial constraints and resulting in structurally faithful 3D scenes. 

However, video diffusion models are typically pretrained to capture temporal dynamics in natural videos~\cite{xing2024survey}, thereby encoding strong inductive biases towards motion continuity and temporal ordering. When directly transferred to cross-view 3D world generation, these biases become misaligned with the inherently spatial and order-invariant nature of multi-view observations. Hence, under large viewpoint disparities between the input and target views, the appearance diffusion model is prone to texture drifting, which degrades overall synthesis quality. To address this challenge, we introduce a simple yet effective random latent shuffle strategy during the training process. By weakening the model’s dependence on view-order–specific cues, this strategy encourages the learning of spatially consistent appearance patterns under varying viewpoints, leading to more robust appearance synthesis and improved cross-view consistency in the generated 3D scenes.

Despite the aforementioned improvements, synthesizing novel views under large  disparities remains a challenging problem. In such cases, the limited spatial coverage of the input image provides weak constraints for distant target viewpoints, causing direct single-pass synthesis to yield implausible geometry. To address this limitation, we propose a test-time scaling strategy that progressively refines the generated scene by incorporating reliable views from earlier synthesis rounds as additional conditioning. Combined with a masked-warping mechanism, this strategy substantially enhances perceptual plausibility without compromising quantitative performance.

Beyond its core design, we further demonstrate experimentally that GTA exhibits strong extensibility and practical versatility. In particular, GTA can be seamlessly integrated as a post-hoc enhancement module to improve the performance of a wide range of existing image-to-3D pipelines, highlighting its compatibility with diverse generation paradigms. Moreover, the proposed framework readily generalizes to related tasks such as 3D scene editing and video depth estimation, underscoring its applicability beyond 3D scene synthesis. Notably, GTA also demonstrates favorable data efficiency, achieving competitive performance with substantially reduced training data, which further enhances its practicality in real-world scenarios.

The main contributions of this paper are summarized as follows:
\begin{itemize}
    \item \textcolor{black}{We propose GTA, a novel image-to-3D world generation framework built upon a two-stage geometry-then-appearance formulation, where multi-view geometry is learned as an explicit generative intermediate and serves as structured guidance for subsequent appearance synthesis.} 

    \item We introduce a simple yet effective random latent shuffle strategy for training video diffusion models in cross-view 3D world generation, which promotes the learning of spatially consistent appearance representations and significantly improves cross-view coherence under large viewpoint variations.

    \item We propose a test-time scaling strategy to further \textcolor{black}{improve synthesis quality under large viewpoint disparities.} By progressively refining the synthesized scene using reliable views from earlier generations, together with a masked-warping mechanism, this strategy substantially improves perceptual plausibility while maintaining comparable quantitative performance.

    \item Extensive experimental results demonstrate that GTA achieves state-of-the-art performance in 3D world generation, as well as exhibiting strong versatility as a general enhancement module and supporting multiple downstream applications with favorable data efficiency during model training.
    
\end{itemize}

\section{Related Works}

\subsection{Video Generation}
Recently, significant progress has been made in video generation~\cite{xing2024survey,waseem2025video,ma2025video}, driven by the rapid development of large-scale datasets and generative models. By extending image generation models to the temporal domain, these approaches explicitly model spatiotemporal dependencies and enable the synthesis of high-quality video sequences with improved visual fidelity and temporal coherence~\cite{tulyakov2018mocogan,li2018video,singer2022make,tian2024videotetris,gupta2024photorealistic,chen2024gentron,wang2025lavie,zeng2024make}.

Building upon this foundation, a growing body of research has extended video generative models to a variety of downstream scenarios, including controllable video generation~\cite{ma2025controllable,hao2018controllable,hu2022make,shi2024motion}, long-range video modeling~\cite{li2024survey,ge2022long,henschel2025streamingt2v,lu2024freelong}, and efficient video synthesis~\cite{zhan2025bidirectional,xing2024simda}. For example,
VideoControlNet~\cite{hu2023videocontrolnet} augments video diffusion models with an explicit control pathway that conditions generation on structural cues, enabling precise spatiotemporal control without retraining the backbone model. MoC~\cite{cai2025mixture} addresses the challenge of extended temporal generation by dynamically combining multiple contextual windows at different temporal scales, allowing the model to capture both short-term dynamics and long-range dependencies for coherent long-video synthesis. To further improve efficiency, BSA~\cite{zhan2025bidirectional} proposes a sparsity-aware attention mechanism that restricts spatiotemporal interactions to structured bidirectional neighborhoods, significantly accelerating video diffusion training while maintaining generation quality.

In light of the remarkable progress of video generative models in modeling complex spatiotemporal dynamics, we observe that multi-view observations of a static 3D scene can be naturally formulated as a structured video sequence indexed by camera viewpoints rather than time. This perspective motivates us to adopt video generative models as a principled foundation for 3D world generation in this work.

\subsection{3D Generation}
3D generation has recently attracted increasing attention due to its broad applicability in areas such as autonomous driving, robotics, and immersive environments~\cite{li2024advances,wen20253d,li2024instant3d,moschoglou20203dfacegan,di2025hyper,song2023automatic}. To this end, existing approaches can be broadly categorized into two paradigms: 3D object generation, which concentrates on synthesizing individual objects~\cite{poole2022dreamfusion,lin2023magic3d,raj2023dreambooth3d,wang2023score}, and 3D world generation, which targets the generation of complex scenes with multiple objects and rich structural layouts~\cite{sun2024dimensionx,bahmani2025ac3d,liang2025wonderland}. We review these two lines of work below.

\paragraph{3D object generation.}
Owing to the absence of large-scale datasets with precise 3D annotations, early approaches to 3D object generation primarily relied on pretrained 2D generative priors and optimization-based frameworks without explicit 3D supervision~\cite{poole2022dreamfusion,lin2023magic3d,wang2023prolificdreamer,xu2023neurallift,tang2023make}. For example, NeuralLift-360~\cite{xu2023neurallift} optimizes a NeRF representation guided by CLIP, enforcing semantic consistency between novel-view renderings and the reference image in a shared feature space, thereby enabling approximate 360\textdegree radiance field reconstruction from a single input image. Make-It-3D~\cite{tang2023make} further incorporates pretrained diffusion models into the optimization process, leveraging their strong generative priors to impose perceptual and semantic constraints on novel-view renderings, which leads to higher-quality 3D asset generation. Despite their strong visual performance, these approaches fundamentally rely on scene-specific optimization and are highly sensitive to the quality of pretrained 2D priors, resulting in substantial computational overhead and limited scalability, which hampers their practical applicability.

Recently, with the emergence of large-scale 3D datasets providing explicit geometric and appearance supervision (e.g., Objaverse~\cite{deitke2023objaverse}, which offers diverse object shapes and textures at scale) research has gradually shifted toward training feed-forward 3D generation models in an end-to-end manner~\cite{xiang2025structured,liu2023zero,liu2023one,liu2024one,long2024wonder3d}. Leveraging explicit 3D supervision, these approaches avoid per-scene optimization and significantly improve efficiency and scalability. For instance, Zero123~\cite{liu2023zero} introduces a latent diffusion–based framework for view-conditioned 3D generation, enabling the synthesis of novel views given a reference image and target camera poses.
Building upon this paradigm, One-2-3-45~\cite{liu2023one} and its enhanced variant One-2-3-45++~\cite{liu2024one} lift the multi-view images generated by Zero123 into 3D space by enforcing multi-view consistency, reconstructing complete textured 3D meshes with substantially reduced inference time. In a similar spirit, Wonder3D~\cite{long2024wonder3d} proposes a cross-domain diffusion framework that jointly models 2D appearance and 3D geometry, enabling single-image 3D generation by aligning diffusion processes across image and 3D domains.

However, these methods are largely object-centric, typically assuming a single foreground object in a simplified environment, and thus are not well suited for modeling complex 3D scenes with multiple objects and rich structural layouts.

\paragraph{3D world generation.}
In contrast to object-centric settings, 3D world generation aims to synthesize large-scale environments with complex geometry, diverse appearances, and long-range consistency, which poses significantly greater challenges~\cite{yu2024wonderjourney,chung2023luciddreamer,yu2025wonderworld,ni2025wonderturbo}. Early approaches such as~\cite{yu2024wonderjourney} and~\cite{chung2023luciddreamer} address this problem by progressively extending views along predefined camera trajectories, leveraging pretrained image inpainting models to hallucinate unseen regions while preserving global scene continuity. Building upon this incremental generation paradigm, subsequent methods such as~\cite{yu2025wonderworld} and~\cite{chung2023luciddreamer} improve generation efficiency by accelerating the scene expansion process, enabling faster construction of coherent large-scale 3D worlds.

Recently, several works have begun to incorporate view-controllable video diffusion models to enhance scene-level modeling~\cite{he2024cameractrl,yu2024viewcrafter,chen2025flexworld,popov2025camctrl3d,ren2025gen3c}. For instance, CamCtrl3D~\cite{popov2025camctrl3d} conditions an image-to-video diffusion model on predefined 3D camera trajectories by explicitly encoding camera parameters into the latent diffusion process, enabling precise camera control and allowing users to freely navigate virtual 3D scenes from a single input image. \textcolor{black}{Meanwhile, several works have also explored geometry-aware scene-level generation by incorporating geometric cues through rendering, structured conditioning, or intermediate 3D representations.}
For example, ViewCrafter~\cite{yu2024viewcrafter} adopts video diffusion models as strong generative priors, where coarse scene geometry is first estimated using methods such as DUSt3R~\cite{wang2024dust3r}, and then leveraged to condition video generation for completing unobserved regions. This design enables the synthesis of high-fidelity novel-view sequences with large viewpoint variations from one or a few input images. Building upon this line of research, methods such as FlexWorld~\cite{chen2025flexworld} and Gen3C~\cite{ren2025gen3c} \textcolor{black}{introduce persistent 3D cache representations that progressively accumulate geometric information during generation.} By iteratively accumulating and reusing intermediate 3D representations, these approaches can gradually construct complete scenes that support 360\textdegree exploration under single-image conditions. \textcolor{black}{Relatedly, DualCamCtrl~\cite{zhang2025dualcamctrl} and Voyager~\cite{huang2025voyager} further advance geometry-aware scene-level generation by jointly predicting RGB and depth in dual-branch diffusion frameworks, where the explicit modeling of appearance and geometry improves structural consistency and enables more accurate 3D world generation.}

\textcolor{black}{However, despite their impressive results, existing methods are typically appearance-centric or depend on the joint generation of geometry and appearance, where structural cues and visual content remain entangled during optimization. In contrast, in this paper we propose GTA, a novel 3D generation framework centered on a learned generative geometry stage. Instead of deriving geometry solely from an external reconstruction module, GTA explicitly generates multi-view geometry as an intermediate representation and then uses it to guide appearance synthesis in a sequential geometry-then-appearance paradigm. This formulation yields a clearer separation between structural modeling and appearance synthesis, reducing interference between the two while improving geometric reliability and appearance consistency.}

\section{Preliminaries}

\subsection{Video Diffusion Models}
Diffusion models~\cite{ho2020denoising,ho2022video} have recently emerged as a powerful class of generative models for high-dimensional data synthesis, including images and videos. A diffusion model consists of a \emph{forward diffusion process} $q$ that gradually corrupts clean data with Gaussian noise, and a \emph{reverse denoising process} $p_\theta$ that learns to invert this corruption process and recover the data distribution.

Given a clean video sample $\mathbf{x}_0 \sim q_0(\mathbf{x}_0)$, the forward process incrementally adds noise over $\mathcal{T}$ timesteps according to
\begin{equation}
\mathbf{x}_t = \alpha_t \mathbf{x}_0 + \sigma_t \boldsymbol{\epsilon}, \quad \boldsymbol{\epsilon} \sim \mathcal{N}(\mathbf{0}, \mathbf{I}),
\end{equation}
where $\alpha_t$ and $\sigma_t$ are predefined noise schedules. The reverse process parameterized by $\theta$ aims to denoise $\mathbf{x}_t$ by predicting the injected noise $\boldsymbol{\epsilon}$ using a neural network $\boldsymbol{\epsilon}_\theta$, which is trained by minimizing the standard noise prediction objective:
\begin{equation}
\min_\theta \ \mathbb{E}_{t \sim \mathcal{U}(0,\mathcal{T}), \boldsymbol{\epsilon} \sim \mathcal{N}(\mathbf{0}, \mathbf{I})}
\left[
\left\| \boldsymbol{\epsilon}_\theta(\mathbf{x}_t, t) - \boldsymbol{\epsilon} \right\|_2^2
\right].
\label{eq:vdm_noise_pred}
\end{equation}

To improve computational efficiency for high-resolution video generation, latent diffusion models (LDMs) are widely adopted~\cite{rombach2022high}. Instead of performing diffusion directly in pixel space, video frames are first encoded into a compact latent representation using a pretrained variational autoencoder (VAE). Both the forward noising process and the reverse denoising process are then carried out in this latent space, significantly reducing computational cost and memory consumption. The final video is obtained by decoding the denoised latent sequence back to the pixel space through the VAE decoder.

By explicitly modeling temporal correlations across frames, video diffusion models capture rich spatiotemporal dynamics and generate temporally coherent video sequences. In this work, we build upon an image-to-video diffusion framework, which aligns naturally with our goal of synthesizing novel views from a single-view input by interpreting multi-view observations as structured video sequences.

\section{Methods}
Given an input image, our goal is to synthesize a consistent 3D world that captures the complexity of real-world scenes. To this end, existing methods typically either emphasize appearance generation without explicitly modeling the underlying geometry~\cite{yu2024viewcrafter,chen2025flexworld,ren2025gen3c,ma2025you}, or jointly predict geometry and appearance within a single framework~\cite{huang2025voyager,zhang2025dualcamctrl,dai2025fantasyworld}. While achieving promising results, the lack of explicit geometric constraints or the tight coupling between geometry and appearance, which are two modalities with distinct learning characteristics, often leads to ambiguities during generation, resulting in unreliable structure estimation and degraded cross-view consistency.

To solve this problem, we propose \textbf{GTA}, a novel method that formulates image-to-3D world generation in a stage-wise manner, explicitly disentangling geometry generation from appearance synthesis (Sec.~\ref{sec 4.1}). To further enhance cross-view appearance consistency, we introduce a random latent shuffle strategy during training (Sec.~\ref{sec 4.2}), together with a test-time scaling strategy that improves scene synthesis quality during inference (Sec.~\ref{sec 4.3}). Details are provided below.

\subsection{Geometry Then Appearance Video Diffusion}~\label{sec 4.1}
Motivated by the coarse-to-fine nature of human visual perception~\cite{navon1977forest,watt1987scanning}, where global scene structure is inferred prior to fine-grained surface details, we posit that scene-level 3D generation should likewise prioritize geometric structure before appearance synthesis. In this context, geometry provides a global and view-consistent spatial scaffold, while appearance encodes higher-frequency visual details that are naturally conditioned on such structure.

Building upon this insight, we design GTA to explicitly follow a hierarchical modeling principle. Specifically, GTA decomposes image-to-3D world generation into two successive stages, each handled by a dedicated video diffusion model. In the first stage, a geometry video diffusion model predicts a multi-view geometric representation of the scene from the input image. In the second stage, an appearance video diffusion model synthesizes fine-grained textures conditioned on the predicted geometry, enabling appearance generation to be guided by explicit structural priors rather than inferred implicitly.

Formally, let $I_0 \in \mathbb{R}^{H \times W \times 3}$ denote the input image captured from a reference viewpoint $v_0$. Our goal is to generate a set of view-consistent observations $\mathcal{O}$ of the 3D world from a collection of $T$ target viewpoints $\{v_t\}_{t=1}^{T}$. Each observation consists of a geometric component and a corresponding appearance component, which we denote as
\begin{equation}
\mathcal{O} = \{ (G_t, A_t) \}_{t=1}^{T},
\end{equation}
where $G_t$ represents the geometric structure of the scene observed from viewpoint $v_t$, and $A_t$ denotes the corresponding RGB appearance. Details of the geometry and appearance modeling are illustrated below.

\begin{figure*}[t]
\centering
\includegraphics[width=1\textwidth]{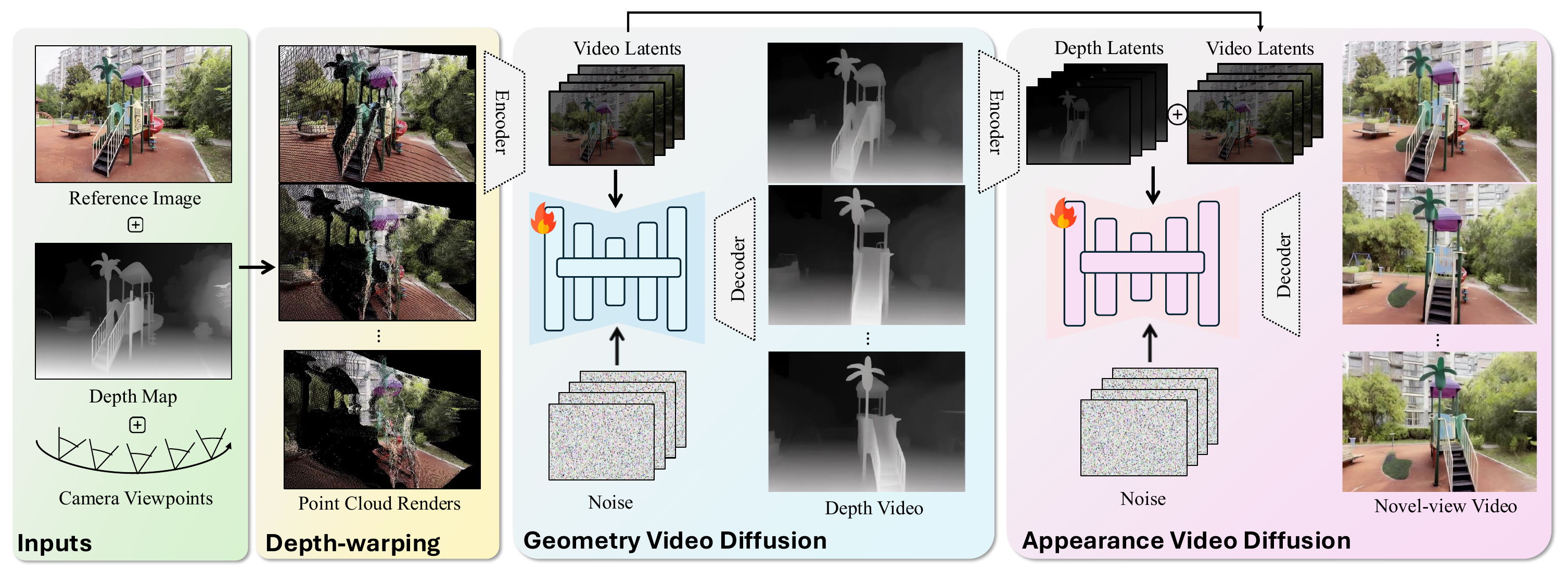}
\caption{\textbf{Pipeline of our proposed geometry then appearance video diffusion.} Starting from a single input image, the method first performs geometry video diffusion to generate a view-consistent multi-view video depth via depth-based warping. Conditioned on the predicted geometry, an appearance video diffusion model then synthesizes a coherent novel-view RGB video with fine-grained textures.}
\label{fig2}
\end{figure*}

\paragraph{Geometry Video Diffusion.}
As shown in Fig.~\ref{fig2}, for the input image $I_0$, we first estimate its depth map $D_0 \in \mathbb{R}^{H \times W \times 1}$ using a pretrained monocular depth estimator. This estimated depth provides an explicit geometric proxy of the scene, allowing us to reason about cross-view spatial correspondence in a principled manner. Leveraging this geometric prior, we construct coarse multi-view observations via depth-based warping, serving as structured conditioning for subsequent geometry modeling.

Specifically, the depth map $D_0$ is first lifted into a set of 3D point clouds $\mathcal{P}$ using the following equation:
\begin{equation}
\mathcal{P} = \left\{ \mathbf{P}_i \in \mathbb{R}^3 \mid 
\mathbf{P}_i = D_0(\mathbf{u}_i)\, \mathbf{K}^{-1} \tilde{\mathbf{u}}_i \right\}, \label{eq. 4.1.2}
\end{equation}
where $i \in \{1,\dots,HW\}$ indexes pixel locations in the reference image, $\mathbf{u}_i \in \mathbb{R}^2$ denotes the corresponding image coordinate, $\tilde{\mathbf{u}}_i$ is its homogeneous representation, and $\mathbf{K}$ is the camera intrinsic matrix. The resulting point set $\mathcal{P}$ is then rendered into each target viewpoint $v_t$, producing a set of warped RGB observations $\{ \hat{I}_t \}_{t=1}^{T}$. Owing to occlusions and the limited spatial coverage of the reference view, these warped observations are inherently partial, forming an incomplete multi-view sequence.

Subsequently, we adopt a video diffusion model to infer a geometrically consistent scene representation from these warped multi-view inputs. Considering that depth provides a compact yet expressive proxy for geometric structure, we represent scene geometry using depth maps and formulate geometry modeling as the synthesis of a complete depth video conditioned on the warped observations.

To this end, the warped RGB sequence $\{ \hat{I}_t \}_{t=1}^{T}$ is first encoded by a video variational autoencoder (VAE) to obtain a sequence of geometry-aware video latents $\mathbf{z}^{\text{rgb}}$. In parallel, the reference-view depth map $D_0$ is encoded by the same VAE to produce a depth latent $\mathbf{z}^{D_0}$ corresponding to the input view. These two latent representations are then concatenated along the channel dimension and fed into the geometry video diffusion model, which performs denoising in the latent space to generate a complete depth video. Finally, the predicted latent sequence is decoded by the VAE decoder to obtain the refined multi-view depth maps. 
Formally, the whole process can be represented as
\begin{equation}
\begin{aligned}
&\{G_t\}_{t=1}^T
=
\mathcal{D}\!\left(
\Phi_{g}\!\left(
\mathbf{z}^{\text{rgb}} \oplus \mathbf{z}^{D_0}
\right)
\right), \\
&\mathbf{z}^{\text{rgb}} = \mathcal{E}(\{\hat{I}_t\}_{t=1}^{T}), \ 
\mathbf{z}^{D_0} = \mathcal{E}(D_0),
\end{aligned}\label{eq. 4.1.3}
\end{equation}
where $\mathcal{E}(\cdot)$ and $\mathcal{D}(\cdot)$ denote the encoder and decoder of a shared video VAE, respectively, $\Phi_g(\cdot)$ denotes the geometry video diffusion model operating in the latent space, $\oplus$ represents channel-wise concatenation.

\paragraph{Appearance Video Diffusion.}
In the second stage, given the generated multi-view depth video $\{G_t\}_{t=1}^{T}$, our goal is to synthesize a view-consistent appearance video $\{A_t\}_{t=1}^{T}$ that faithfully captures the surface textures and visual details. Compared to geometry generation, appearance synthesis involves higher-frequency information and is inherently more ambiguous, especially under large viewpoint changes. We therefore condition appearance modeling explicitly on the predicted geometry, allowing appearance generation to be guided by reliable structural cues.

Specifically, the generated multi-view depth video $\{G_t\}_{t=1}^{T}$ is first encoded by the same video VAE to obtain a sequence of geometry latents $\mathbf{z}^{\text{geo}}$. The geometry latents are then concatenated with the partial RGB latents $\mathbf{z}^{\text{rgb}}$ obtained in the previous stage along the channel dimension, and passed through a lightweight projection module to match the input dimensionality required by the diffusion backbone. To further strengthen geometric conditioning and encourage structure-aligned appearance synthesis, the projected latent is fused with the geometry latent via a residual addition before being fed into the appearance video diffusion model. This design biases the diffusion process toward generating textures that are consistent with the underlying geometric structure, while preserving sufficient flexibility for modeling high-frequency appearance details.
The whole process is denoted as follows:
\begin{equation}
\begin{aligned}
&\{A_t\}_{t=1}^{T}
=
\mathcal{D}\!\left(
\Phi_{a}(\mathbf{h}^{a})
\right), \\
&\mathbf{h}^{a}
=
\Pi\!\left(
\mathbf{z}^{\text{rgb}} \oplus \mathbf{z}^{\text{geo}}
\right)
+
\mathbf{z}^{\text{geo}}, \\
&\mathbf{z}^{\text{geo}}
=
\mathcal{E}(\{G_t\}_{t=1}^{T}),
\label{eq: appearance video diffusion}
\end{aligned}
\end{equation} 
\textcolor{black}{where $\Pi(\cdot)$ denotes the lightweight projection module with four 3D convolutional layers, $\Phi_a(\cdot)$ denotes the appearance video diffusion model, and $\oplus$ denotes channel-wise concatenation.}

\begin{figure*}[t]
\centering
\includegraphics[width=1\textwidth]{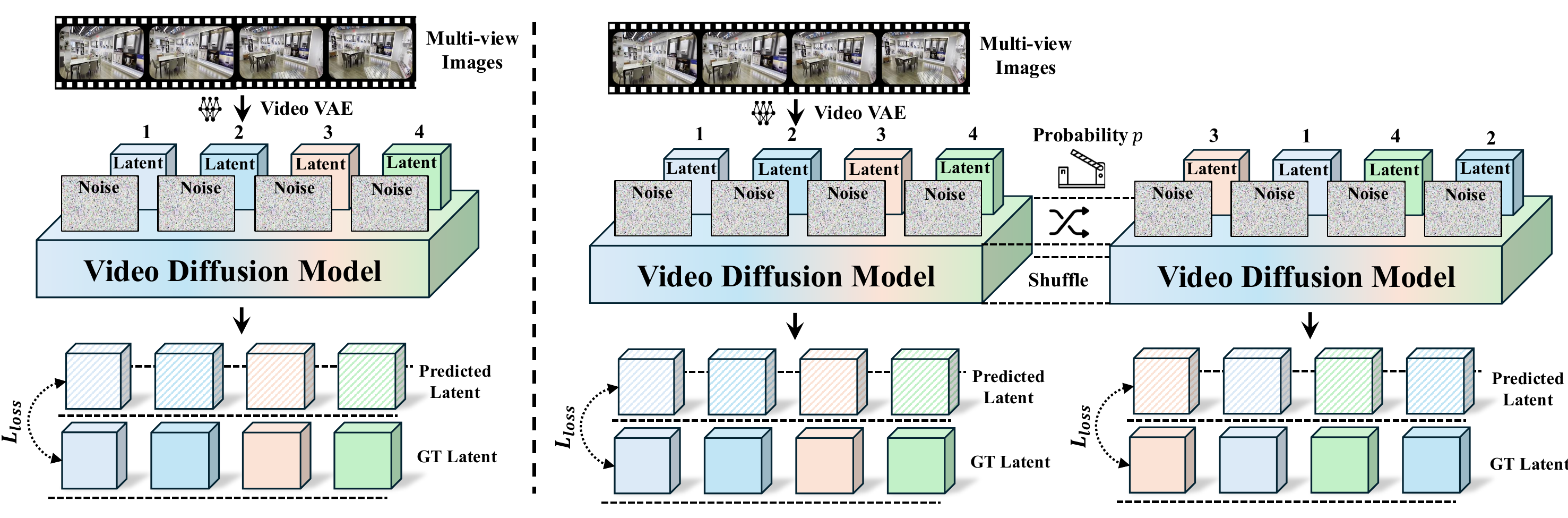}
\caption{\textbf{Pipeline of our proposed random latent shuffle strategy.} During training, the latent representations corresponding to different target views are randomly permuted with probability $p$ before being fed into the video diffusion model. The same permutation is applied to both predicted and ground-truth latents, preserving supervision while discouraging view-order–dependent correlations. This strategy encourages the model to treat multi-view observations as unordered spatial samples rather than temporally ordered frames, and is only applied during training.}
\label{fig3}
\end{figure*}

\subsection{Random Latent Shuffle}~\label{sec 4.2}
After defining the architecture of our geometry then appearance video diffusion models, we now consider how to effectively train it for 3D world synthesis. Intuitively, one may directly adopt the standard training protocol of video diffusion models, treating multi-view observations as ordered video sequences and optimizing the model to predict denoised target views. However, we find that such a straightforward adaptation often leads to noticeable cross-view inconsistencies in practice. For example, regions that are black in the input view may gradually fade into gray as the diffusion process synthesizes a novel target view, particularly under large viewpoint disparities between the input and target views, as shown in Fig.~\ref{fig8}(a).

We attribute this behavior primarily to the inductive biases inherited from video diffusion pretraining. Specifically, video diffusion models are typically optimized to model temporal dynamics in natural videos, where appearance evolution is strongly coupled with smooth motion and a well-defined temporal order. As a result, the learned representations implicitly favor continuity along the sequence dimension and encourage gradual appearance transitions between adjacent frames.

When such models are directly applied to cross-view 3D world synthesis, these temporal assumptions become suboptimal. Multi-view observations of a static scene are inherently spatial in nature and invariant to view ordering, yet the model continues to interpret view indices as a proxy for temporal progression. This misalignment becomes particularly problematic under large viewpoint changes, where enforcing temporal smoothness across views introduces spurious appearance correlations. Consequently, the diffusion process may gradually alter texture statistics across views, leading to visible drifting artifacts and degraded cross-view consistency in the synthesized results.

To solve this problem, we propose a simple yet effective training strategy, termed \textit{Random Latent Shuffle}. The core idea of this strategy is to attenuate the model’s dependence on view-order–specific correlations, while largely preserving the strong generative priors acquired through video diffusion pretraining. Concretely, as shown in Fig.~\ref{fig3}, during training, we stochastically permute the ordering of latent representations associated with different target views with a predefined probability before they are processed by the video diffusion model, which is formulated as:
\begin{equation}
\tilde{\mathbf{z}} = \boldsymbol{\pi}\!\left(\mathbf{z}\right), 
\boldsymbol{\pi} \sim
\begin{cases}
\mathcal{U}(\mathcal{S}_T), & \text{with probability } p, \\
\mathrm{Id}, & \text{with probability } 1-p,
\end{cases}
\end{equation}
where $\mathbf{z}$ denotes the latent representations, and $\tilde{\mathbf{z}}$ is the corresponding latent after random shuffling. The operator $\boldsymbol{\pi}$ represents a permutation applied along the view dimension. With probability $p$, $\boldsymbol{\pi}$ is sampled uniformly from the permutation group $\mathcal{S}_T$, denoted by $\mathcal{U}(\mathcal{S}_T)$, which randomly reorders the $T$ target views; otherwise, the identity permutation $\mathrm{Id}$ is applied, leaving the latent ordering unchanged. The probability $p$ controls the frequency of applying random latent shuffle during training.

\textcolor{black}{During training, the same permutation $\boldsymbol{\pi}$ is applied to the latent to be denoised, the corresponding ground-truth latent, and the associated warped conditioning latent, ensuring that all view-dependent supervision and conditioning signals remain aligned after shuffling.} The video diffusion model is then optimized using the standard noise prediction objective, following the conventional training protocol. In this way, random latent shuffle perturbs the view ordering without altering the underlying supervision signal, effectively discouraging the model from exploiting spurious order-dependent correlations.

Notably, the random latent shuffle is only employed as a training-time regularization strategy. At inference time, no shuffling is applied, and the latent representations are fed into the video diffusion model in the target view order. 

\begin{figure}[t]
\centering
\includegraphics[width=0.49\textwidth]{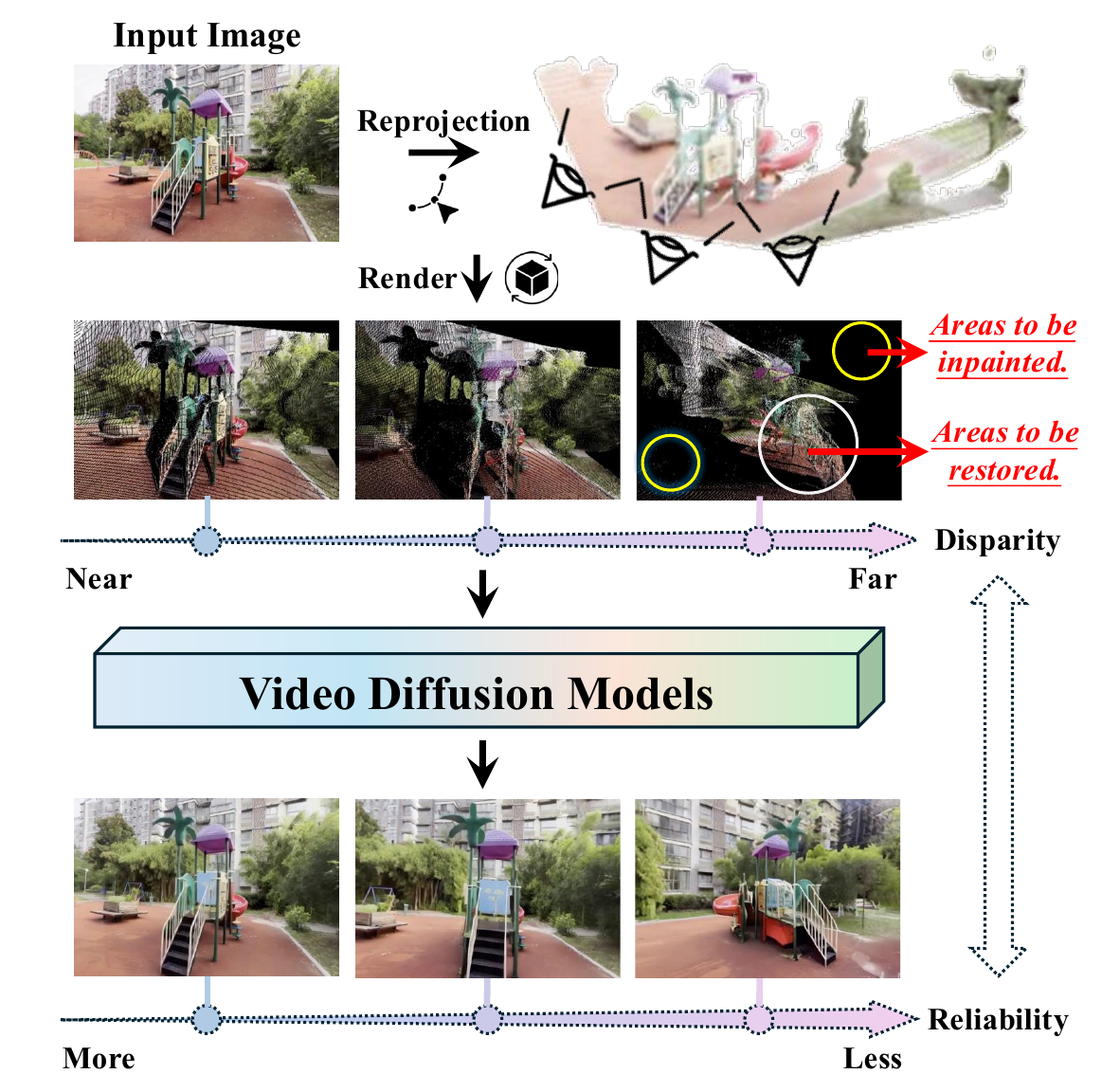}
\caption{\textbf{Two key observations from single-pass inference.}
The reliability of synthesized views decreases as the viewpoint disparity from the input increases. Meanwhile, the rendered partial observations contain two distinct region types: visible regions requiring appearance restoration, and unobserved regions that must be hallucinated via inpainting.}
\label{fig4}
\end{figure}

\begin{figure*}[t]
\centering
\includegraphics[width=0.99\textwidth]{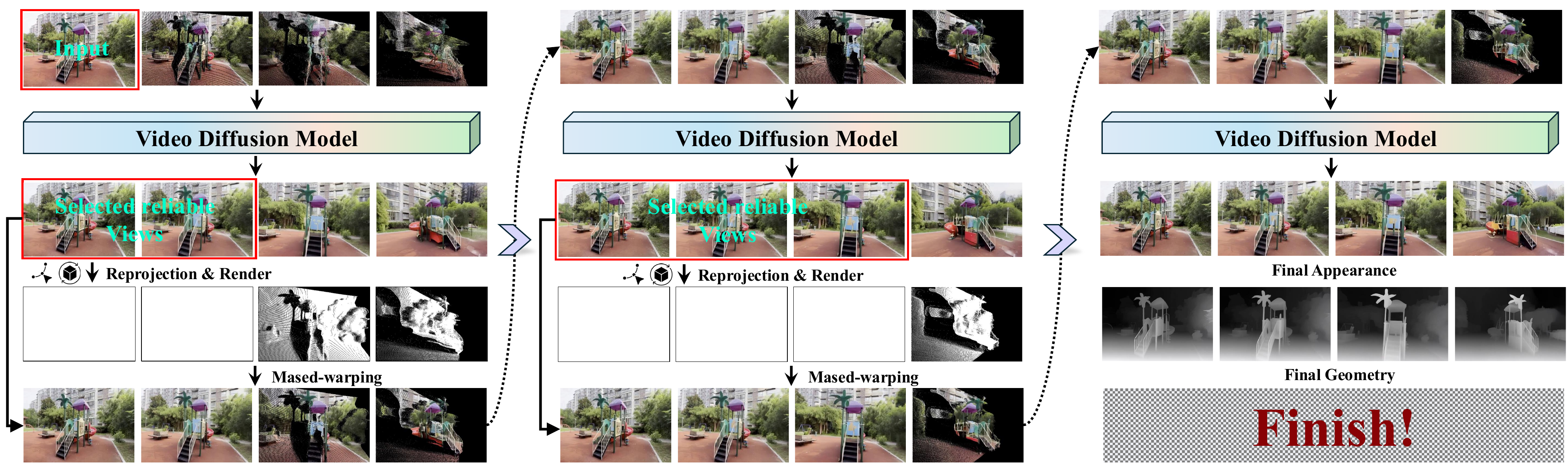}
\caption{\textbf{Pipeline of the proposed test-time scaling strategy.}
Starting from an initial inference pass, a subset of reliable views is progressively selected and reprojected to construct updated partial observations. Through masked-warping, restored regions are fixed while unreliable regions are suppressed, and the video diffusion model is iteratively conditioned on the refined inputs, enabling coarse-to-fine synthesis of distant target views with improved stability and cross-view consistency.}
\label{fig5}
\end{figure*}

\subsection{Test-time Scaling}~\label{sec 4.3}
Although the proposed random latent shuffle strategy substantially enhances cross-view consistency, synthesizing high-fidelity novel views under large viewpoint disparities remains challenging at inference time. In such cases, the limited spatial coverage provided by the input observation imposes weak geometric and appearance constraints on distant target views, often leading to implausible structures or visually inconsistent results.

To address this limitation, we pursue a test-time scaling strategy that enhances synthesis quality by more effectively leveraging the model’s intrinsic generative capacity and test-time information, without incurring additional training overhead. To this end, as shown in Fig.~\ref{fig4}, we begin by making two key observations from a single-pass inference setting.

First, the reliability of synthesized novel views exhibits a strong dependence on their distance to the input view. \textbf{\textit{Views with relatively small disparities tend to be generated more faithfully,}} as abundant geometric and appearance cues can be effectively transferred from the input image. In contrast, views that lie farther are constrained by substantially weaker visual evidence and are therefore more susceptible to structural inaccuracies and visual artifacts, making their rendering inherently less reliable.

Second, in our pipeline, where a partial RGB video sequence is rendered from point-based representations and subsequently completed by video diffusion models to synthesize a full multi-view RGB sequence, \textbf{\textit{the models are actually required to handle two fundamentally different types of generation simultaneously.}} On the one hand, for regions that are visible from the input view under the target viewpoints, the rendered sequence provides partial but imperfect information, typically affected by distortions or rendering artifacts. These regions mainly demand appearance correction and refinement rather than content hallucination, and are therefore referred to as \textit{areas to be restored}. On the other hand, for regions that are not visible from the input views under the target viewpoints, no reliable visual evidence is available, and the models must rely on their learned generative priors to plausibly synthesize the missing content. We refer to these regions as \textit{areas to be inpainted.}

\begin{algorithm}[t]
\caption{Test-time Scaling}\label{alg 1}
\label{alg:test-time-scaling}
\begin{algorithmic}[1]

\State \textbf{Input:} input image $I$, target viewpoints $\{v_t\}_{t=1}^{T}$, trained video diffusion models
\State \textbf{Output:} synthesized RGB sequence $\{A_t\}_{t=1}^{T}$

\State Perform an initial inference pass using Eq.~\ref{eq. 4.1.3} and Eq.~\ref{eq: appearance video diffusion} to obtain $\{A_t\}_{t=1}^{T}$
\State Initialize the reliable view set $\mathcal{R} \leftarrow \{1,\ldots,Q\}$

\While{$\mathcal{R}$ does not cover all target views}
    \State Perform depth-based warping using Eq.~\ref{eq. 4.1.2} with $I$ and $\{A_t\}_{t\in\mathcal{R}}$
    \State Derive a binary visibility mask sequence $\{\hat{M}_t\}_{t=1}^{T}$
    \State Apply masked-warping to obtain a masked partial RGB sequence $\{\hat{I}_t\}_{t=1}^{T}$
    \State Update $\{A_t\}_{t=1}^{T}$ via video diffusion conditioned on $\{\hat{I}_t\}_{t=1}^{T}$
    \State Expand $\mathcal{R}$ to include additional $Q$ target views
\EndWhile

\State \Return $\{A_t\}_{t=1}^{T}$
\end{algorithmic}
\end{algorithm}

Based on the above two observations, as illustrated in Fig.~\ref{fig5} and Alg.~\ref{alg 1}, our test-time scaling proceeds as follows. Without loss of generality, we assume that the target views are ordered by increasing viewpoint distance from the input view. After an initial inference pass with Eq.~\ref{eq. 4.1.3} and Eq.~\ref{eq: appearance video diffusion}, the first $Q$ views in the generated RGB sequence $\{A_t\}_{t=1}^{T}$ are regarded as reliable. Together with the input view, they are then used to perform depth-based warping using Eq.~\ref{eq. 4.1.2} along the same $T$ target viewpoints $\{v_t\}_{t=1}^{T}$, yielding an updated partial observations.

From these warped results, we further derive a binary mask sequence $\{ \hat{M}_t \}_{t=1}^{T}$ that indicates pixel-wise visibility, where $\hat{M}_t(\mathbf{x}) = 1$ denotes pixels that can be rendered from the available views, and $\hat{M}_t(\mathbf{x}) = 0$ corresponds to regions that remain unobserved. Following the second observation, regions corresponding to \textit{areas to be restored} are expected to be largely reliable at this stage, as coarse but informative cues can already be inferred from the input view and are reasonably approximated after an initial forward pass of the model. We therefore apply the mask sequence to the synthesized RGB outputs $\{A_t\}_{t=1}^{T}$, retaining the restored regions while suppressing unreliable content in the \textit{areas to be inpainted}, thereby producing a masked partial RGB sequence $\{ \hat{I}_t \}_{t=1}^{T}$. We refer to this operation as masked-warping. By fixing restored regions, the models are freed from jointly performing restoration and hallucination, effectively reducing interference between the two fundamentally different modes and yielding more stable synthesis under large disparities.

The resulting masked partial RGB sequence is then fed back into the video diffusion models as input for the next inference round. This procedure is performed iteratively in a progressive manner, where an increasing set of synthesized views is treated as reliable at each iteration. At every stage, the currently reliable views are used to perform depth-based warping and masked-warping to construct an updated partial RGB sequence, which serves as conditioning for subsequent inference. The process continues until all target views are incorporated or a predefined number of iterations is reached.

In essence, this test-time scaling strategy facilitates a coarse-to-fine inference paradigm, where reliable visual evidence is progressively accumulated and leveraged to guide subsequent synthesis. As a result, distant target views benefit from increasingly stronger conditioning, leading to more stable and visually consistent novel view generation under large disparities, while incurring no additional training overhead.

\section{Experiments}

\subsection{Experimental Setups}
\noindent\textbf{\textit{Datasets.}} 
We conduct training on the DL3DV dataset~\cite{ling2024dl3dv}, which comprises approximately 10K real-world scenes captured from diverse camera viewpoints. For each scene, we curate 10 high-quality video sequences featuring varying camera trajectories, resulting in a total of 100K training samples. 

At inference time, we evaluate our model on two datasets. The first is the official DL3DV evaluation split~\cite{ling2024dl3dv}, which contains 55 scenes that are strictly disjoint from the training set. For this split, we construct 55 videos, each composed of a sequence of multi-view images. Given the first frame of each video as input, the model is required to generate the remaining frames. In addition, we assess the generalization ability of our method on a fully held-out dataset, RealEstate10K~\cite{zhou2018stereo}. We randomly sample 100 scenes from the dataset and remove those containing insufficient multi-view images, resulting in 79 valid scenes. Following the same protocol as DL3DV, we construct 79 multi-view videos and perform inference on this set.

\vspace{4mm}
\noindent\textbf{\textit{Baseline and Metrics.}}
We compare our method with several representative state-of-the-art approaches spanning two fundamental paradigms of 3D scene generation. Specifically, we compare against two categories of approaches: (i) \emph{\textbf{appearance-only generation}} methods that focus solely on synthesizing visual appearance (e.g., texture and color) without explicitly modeling scene geometry, and (ii) methods that \textit{\textbf{simultaneously generate appearance and geometry}} by jointly reasoning about visual content and structural information. For the former, we compare with See3D~\cite{ma2025you}, ViewCrafter~\cite{yu2024viewcrafter}, FlexWorld~\cite{chen2025flexworld}, Gen3C~\cite{ren2025gen3c}, and TrajectoryCrafter~\cite{yu2025trajectorycrafter}, covering both U-Net--based methods (See3D and ViewCrafter) and Diffusion Transformer (DiT)--based methods (FlexWorld, Gen3C, and TrajectoryCrafter); for the latter, we include Voyager~\cite{huang2025voyager} as a representative baseline.

We evaluate the performance of all methods from three complementary perspectives. First, to assess \emph{\textbf{fidelity}}, we adopt widely used pixel- and perceptual-level metrics, including PSNR, SSIM, and LPIPS, which respectively measure reconstruction accuracy, structural similarity, and perceptual distance. Second, to evaluate overall \emph{\textbf{visual quality}}, we employ FID and Q-Align~\cite{wu2023q}. 
FID measures the distribution-level consistency between generated and real videos, while Q-Align provides fine-grained, human-aligned quality assessment across multiple perceptual dimensions. Specifically, Q-Align evaluates both video-level and frame-level perceptual quality and aesthetics, including Q-Align-V (quality), Q-Align-V (aesthetics), Q-Align-I (quality), and Q-Align-I (aesthetics). All Q-Align scores are normalized to a range of $[0,5]$, where higher values indicate better alignment with human visual perception. \textcolor{black}{Finally, following the evaluation protocol of CameraCtrl~\cite{he2024cameractrl}, we assess \emph{\textbf{geometric accuracy}} by measuring the deviation between the estimated camera trajectories of generated videos and the corresponding reference trajectories. Specifically, for each generated video, we use the official pretrained VGGT model~\cite{wang2025vggt} to estimate the camera pose of each frame. The reference trajectory is obtained by applying the same VGGT pipeline to the corresponding ground-truth video. Both trajectories are then converted into relative poses with respect to the first frame before computing rotation error (R-err) and translation error (T-err). R-err reflects discrepancies between rotation matrices, while T-err measures differences in camera translation vectors.}

\vspace{4mm}
\noindent\textbf{\textit{Implementation Details.}}
Our model is obtained via supervised fine-tuning (SFT) based on CogVideo-X-5B~\cite{yang2024cogvideox}, and generates videos consisting of 49 frames at a spatial resolution of $720 \times 480$. During training, camera poses are estimated using VGGT~\cite{wang2025vggt}, while per-frame depth maps are obtained using Video Depth Anything~\cite{chen2025video}. At inference time, depth estimation for the input image is performed using Depth Anything V2~\cite{yang2024depth}. For the random latent shuffle strategy, we set the shuffle probability $p$ to $0.5$. For the test-time scaling, the reliable view window size $Q$ is set to $5$. All training experiments are conducted on 8 NVIDIA A800 GPUs, where the appearance video diffusion and geometry video diffusion models are each trained for 40,000 iterations. During inference, all experiments are performed on a single NVIDIA A800 GPU, with peak GPU memory consumption of approximately 20\,GB.

\begin{table*}[t]
\centering
\small
\setlength{\tabcolsep}{5.5pt}
\resizebox{0.99\textwidth}{!}{%
\begin{tabular}{lccccccc}
\toprule
\textbf{Metrics} 
& \textbf{See3D} 
& \textbf{ViewCrater} 
& \textbf{TrajectoryCrafter} 
& \textbf{FlexWorld} 
& \textbf{Gen3C} 
& \textbf{Voyager} 
& \textbf{Ours} \\
\midrule

\multicolumn{8}{c}{\textbf{DL3DV Dataset}} \\
\midrule
PSNR $\uparrow$ 
& \cellcolor{yellow!25}15.86 & 15.09 & 15.82 & 15.39 & \cellcolor{orange!25}16.15 & 15.23 & \cellcolor{red!25}\textbf{17.47} \\
SSIM $\uparrow$ 
& \cellcolor{orange!25}0.483 & 0.409 & 0.454 & 0.428 & \cellcolor{yellow!25}0.473 & 0.406 & \cellcolor{red!25}\textbf{0.525} \\
LPIPS $\downarrow$ 
& 0.425 & 0.457 & \cellcolor{yellow!25}0.412 & \cellcolor{orange!25}0.402 & 0.433 & 0.436 & \cellcolor{red!25}\textbf{0.342} \\
\midrule
FID $\downarrow$ 
& 65.70 & 61.15 & \cellcolor{yellow!25}43.94 & \cellcolor{orange!25}39.53 & \cellcolor{yellow!25}43.56 & 60.90 & \cellcolor{red!25}\textbf{38.32} \\
Q-Align-V (quality) $\uparrow$ 
& 3.045 & 3.066 & \cellcolor{orange!25}3.361 & \cellcolor{yellow!25}3.312 & 2.954 & 3.305 & \cellcolor{red!25}\textbf{3.385} \\
Q-Align-V (aesthetics) $\uparrow$ 
& 2.258 & 2.224 & \cellcolor{orange!25}2.392 & 2.280 & 2.097 & \cellcolor{yellow!25}2.292 & \cellcolor{red!25}\textbf{2.421} \\
Q-Align-I (quality) $\uparrow$ 
& 2.887 & 2.787 & \cellcolor{yellow!25}3.088 & 3.060 & 2.813 & \cellcolor{orange!25}3.209 & \cellcolor{red!25}\textbf{3.215} \\
Q-Align-I (aesthetics) $\uparrow$ 
& 1.962 & 1.927 & \cellcolor{yellow!25}2.023 & 1.901 & 1.817 & \cellcolor{orange!25}2.041 & \cellcolor{red!25}\textbf{2.082} \\
\midrule
T-err $\downarrow$ 
& \cellcolor{orange!25}0.003 & 0.008 & \cellcolor{yellow!25}0.005 & \cellcolor{red!25}\textbf{0.001} & \cellcolor{orange!25}0.003 & 0.008 & \cellcolor{red!25}\textbf{0.001} \\
R-err $\downarrow$ 
& \cellcolor{orange!25}0.017 & 0.029 & 0.026 & \cellcolor{yellow!25}0.019 & 0.022 & 0.025 & \cellcolor{red!25}\textbf{0.013} \\

\midrule
\midrule
\multicolumn{8}{c}{\textbf{RealEstate Dataset}} \\
\midrule
PSNR $\uparrow$ 
& 15.97 & 13.52 & \cellcolor{yellow!25}16.08 & 15.77 & \cellcolor{orange!25}16.46 & 15.80 & \cellcolor{red!25}\textbf{17.01} \\
SSIM $\uparrow$ 
& 0.595 & 0.472 & 0.589 & \cellcolor{yellow!25}0.597 & \cellcolor{orange!25}0.612 & 0.574 & \cellcolor{red!25}\textbf{0.644} \\
LPIPS $\downarrow$ 
& 0.380 & 0.485 & \cellcolor{yellow!25}0.376 & 0.380 & \cellcolor{orange!25}0.368 & 0.398 & \cellcolor{red!25}\textbf{0.341} \\
\midrule
FID $\downarrow$ 
& 44.99 & 50.37 & 32.36 & \cellcolor{orange!25}27.48 & \cellcolor{yellow!25}31.08 & 37.08 & \cellcolor{red!25}\textbf{26.72} \\
Q-Align-V (quality) $\uparrow$ 
& 2.877 & 2.608 & \cellcolor{yellow!25}3.237 & 3.198 & 2.889 & \cellcolor{orange!25}3.260 & \cellcolor{red!25}\textbf{3.455} \\
Q-Align-V (aesthetics) $\uparrow$ 
& 1.980 & 1.796 & \cellcolor{orange!25}2.236 & 2.127 & 1.968 & \cellcolor{yellow!25}2.229 & \cellcolor{red!25}\textbf{2.417} \\
Q-Align-I (quality) $\uparrow$ 
& 2.623 & 2.474 & \cellcolor{orange!25}3.198 & 2.956 & 2.731 & \cellcolor{yellow!25}3.063 & \cellcolor{red!25}\textbf{3.287} \\
Q-Align-I (aesthetics) $\uparrow$ 
& 1.609 & 1.535 & \cellcolor{orange!25}1.946 & 1.751 & 1.665 & \cellcolor{yellow!25}1.898 & \cellcolor{red!25}\textbf{2.002} \\
\midrule
T-err $\downarrow$ 
& 0.038 & 0.103 & \cellcolor{yellow!25}0.032 & \cellcolor{red!25}\textbf{0.024} & 0.034 & 0.035 & \cellcolor{orange!25}0.026 \\
R-err $\downarrow$ 
& \cellcolor{yellow!25}0.065 & 0.129 & 0.080 & \cellcolor{orange!25}0.060 & 0.075 & 0.094 & \cellcolor{red!25}\textbf{0.041} \\

\bottomrule
\end{tabular}%
}
\caption{\textbf{Quantitative comparison across fidelity, perceptual quality, and geometric accuracy on the DL3DV and RealEstate datasets. }
$\uparrow$ and $\downarrow$ indicate that higher and lower values correspond to better performance, respectively. 
Best, second-best, and third-best results are highlighted in red, orange, and yellow. Our method consistently achieves superior performance on both datasets. Notably, all results here are reported using single-pass inference without test-time scaling, and the effect of test-time scaling is analyzed separately in the ablation study.}
\label{tab:quantitative_comparison_datasets}
\end{table*}

\begin{figure*}[h]
\centering
\includegraphics[width=0.93\textwidth]{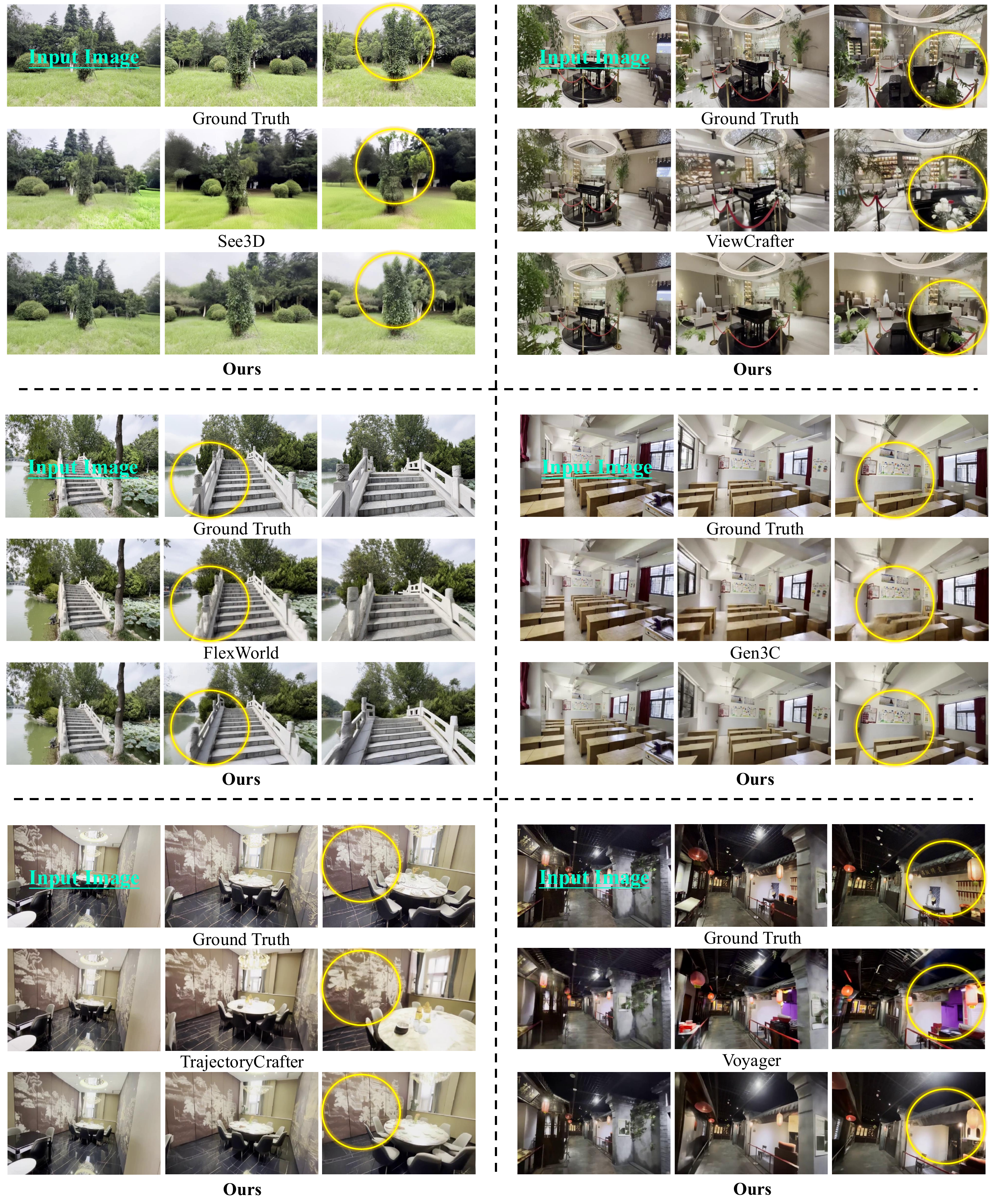}
\caption{\textbf{Qualitative comparison between our method and state-of-the-art approaches.} Our method generates visually coherent scenes with improved detail preservation and geometric consistency, as highlighted in yellow circles.}
\label{fig6}
\end{figure*}

\textcolor{black}{For the projection module $\Pi$ used in the appearance video diffusion model (Eq.~\ref{eq: appearance video diffusion}), we employ a lightweight projector composed of four 3D convolutional layers to align the channel dimensionality of the concatenated RGB and geometry latents before feeding them into the appearance diffusion backbone. Specifically, $\Pi$ maps a 32-channel fused latent to 16 channels through four Conv3D layers with channel dimensions $32 \rightarrow 32 \rightarrow 32 \rightarrow 32 \rightarrow 16$, namely
$\mathrm{Conv3D}(32,32,k=3,s=1,p=1)$,
$\mathrm{Conv3D}(32,32,k=3,s=1,p=1)$,
$\mathrm{Conv3D}(32,32,k=3,s=1,p=1)$, and
$\mathrm{Conv3D}(32,16,k=3,s=1,p=1)$.
Here, $k=3$ denotes a $3 \times 3 \times 3$ spatio-temporal kernel, and the stride and padding are applied along the temporal, height, and width dimensions, thereby preserving the spatio-temporal resolution of the latent representation. We apply a LeakyReLU activation between consecutive Conv3D layers. Overall, this design introduces only approximately $0.1$M parameters, making $\Pi$ a lightweight conditioning module in practice.
}

\textcolor{black}{For latent encoding, we reuse the pretrained video VAE from CogVideoX for both RGB frames and depth maps, and keep it frozen during both training and inference. To encode depth with this RGB-trained VAE, we first convert depth maps into disparity and normalize the disparity sequence to $[0,1]$. The normalized single-channel disparity maps are replicated into three channels and then fed into the VAE using the same preprocessing pipeline as RGB frames. After VAE decoding, the three output channels are averaged to obtain a single-channel normalized disparity map, which is then converted back to the required depth representation for supervision, rendering, and evaluation. This design is in line with recent diffusion-based video depth estimation methods such as DepthCrafter~\cite{hu2025depthcrafter}, where pretrained video VAEs are used as practical tokenizers for depth-like signals. In our framework, sharing a frozen VAE also keeps geometry and appearance representations in a common latent space, making the generated geometry latents directly compatible with the pretrained video diffusion backbone.}

\subsection{Main Results}
Tab.~\ref{tab:quantitative_comparison_datasets} reports the quantitative comparison on the DL3DV and RealEstate datasets. Overall, our method consistently outperforms all competing approaches across fidelity, perceptual quality, and geometric accuracy on both datasets, demonstrating its effectiveness and robustness across different generation paradigms and evaluation settings. 

On the DL3DV dataset, our method achieves the best performance on all fidelity metrics, obtaining the highest PSNR and SSIM as well as the lowest LPIPS, indicating superior reconstruction accuracy and perceptual similarity. In terms of visual quality, our method also yields the lowest FID and consistently achieves the highest scores across all Q-Align dimensions, including both video-level and frame-level quality and aesthetics, suggesting improved alignment with human visual perception. Furthermore, our method demonstrates strong geometric consistency. We achieve the lowest rotation error (R-err) and jointly lowest translation error (T-err), highlighting more accurate camera trajectory estimation and improved geometric alignment with the ground-truth scene structure.

On the RealEstate dataset, which is fully held out from training for our method but used during training by other methods, our approach still maintains clear advantages. Specifically, we obtain the best results on PSNR, SSIM, LPIPS, and FID, and consistently outperform all counterparts across all Q-Align metrics. In addition, our method achieves the lowest R-err and competitive T-err, further validating its strong generalization capability.

In addition to quantitative evaluation, we further provide qualitative comparisons in Fig.~\ref{fig6} to visually assess the perceptual differences among methods. For See3D, the generated results tend to exhibit noticeable loss of fine-grained details, where high-frequency textures are over-smoothed, leading to blurred appearances in fine structures and reduced visual sharpness. For other appearance-only methods, such as ViewCrafter, FlexWorld, Gen3C, and TrajectoryCrafter, synthesis under large viewpoint changes is often accompanied by geometric distortions and texture inconsistency, manifested as deformed object structures, spatial misalignment of structural elements, and view-dependent texture variations, ultimately leading to reduced geometric coherence and visual consistency. For voyager, the synthesized results often suffer from excessive color saturation and unnatural color shifts, along with noticeable discrepancies in fine-grained details relative to the input image. In contrast, our proposed method effectively preserves fine-grained visual details and enforces consistent geometry and texture appearance, resulting in 3D scenes that remain visually coherent and faithful to the input image under large viewpoint changes.

\subsection{Ablation Studies}
To demonstrate the effectiveness of our proposed designs, we conduct ablation studies on geometry-then-appearance generation, Random Latent Shuffle, test-time scaling, \textcolor{black}{the projection module in appearance diffusion, and depth-to-VAE encoding strategies. We also provide controlled and complementary analyses on training-data effects and COLMAP-based pose evaluation to further verify comparison fairness and geometric consistency.}

\subsubsection{Geometry Then Appearance Generation.}
To evaluate the effectiveness of the proposed geometry-then-appearance generation strategy, we randomly select 20 DL3DV scenes with large camera viewpoint changes and conduct an ablation comparison against Voyager, which adopts a simultaneous geometry-and-appearance prediction paradigm.

\begin{table}[t]
\centering
\small
\setlength{\tabcolsep}{4.5pt}
\resizebox{\columnwidth}{!}{%
\begin{tabular}{lccc}
\toprule
\textbf{Generation Strategy} & \textbf{PSNR} $\uparrow$ & \textbf{SSIM} $\uparrow$ & \textbf{LPIPS} $\downarrow$ \\
\midrule
SAG
& 15.03 & 0.391 & 0.471 \\
GTA \textbf{(Ours)}
& \cellcolor{red!25}\textbf{17.14} & \cellcolor{red!25}\textbf{0.502} & \cellcolor{red!25}\textbf{0.367} \\
\bottomrule
\end{tabular}%
}
\caption{\textbf{Quantitative ablation study on generation strategies using scenes with large viewpoint changes.} SAG denotes simultaneous appearance and geometry generation, while GTA denotes geometry-then-appearance generation. GTA consistently improves reconstruction fidelity compared to SAG.}
\label{tab2}
\end{table}

\begin{figure}[t]
\centering
\includegraphics[width=0.5\textwidth]{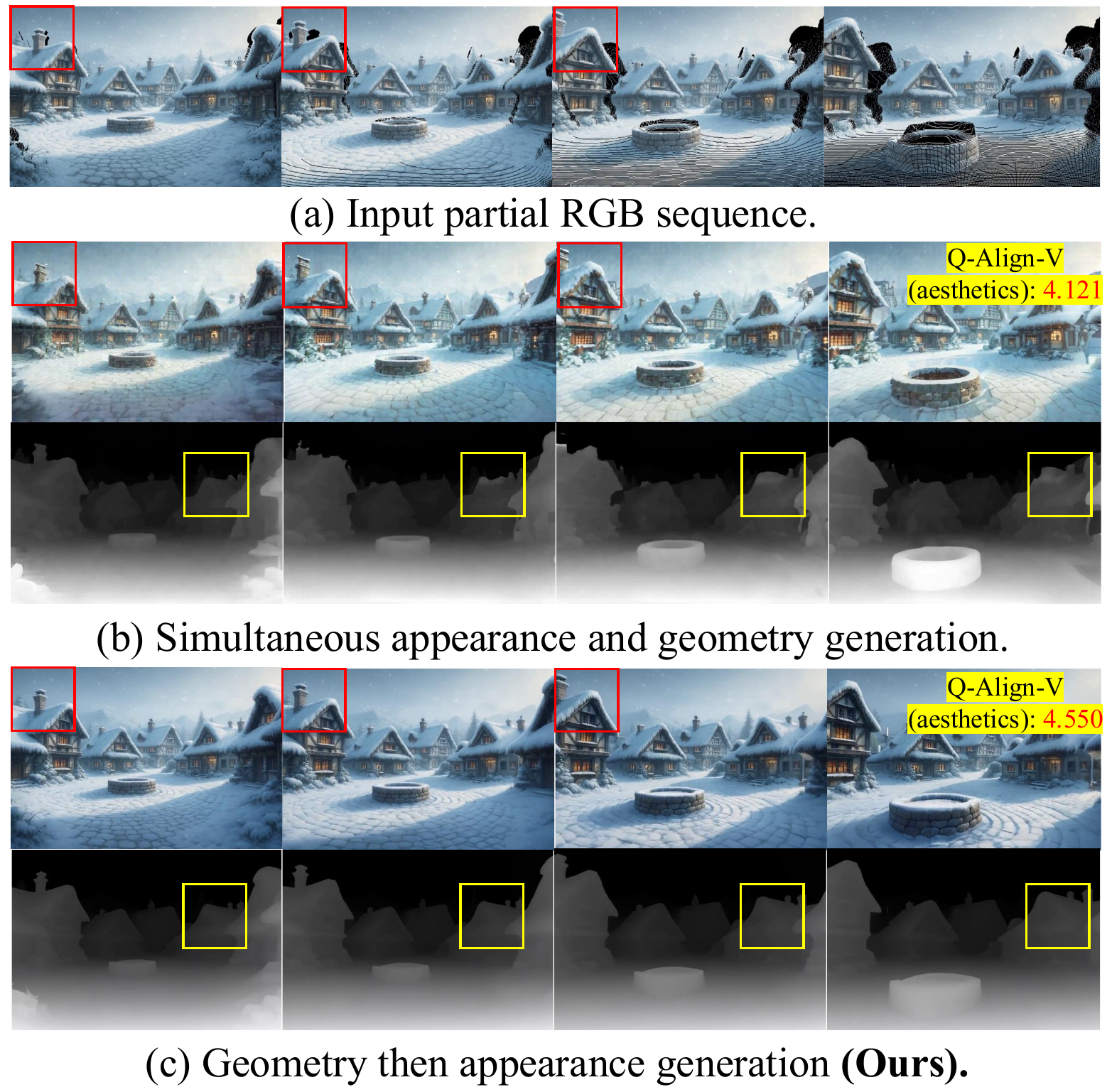}
\caption{\textbf{Qualitative ablation study on geometry-then-appearance generation (GTA). }Compared to simultaneous appearance-and-geometry generation, GTA produces more consistent textures and stable geometry. The red and yellow boxes highlight representative regions where the contrast between the two approaches is particularly evident.}
\label{fig7}
\end{figure}

As demonstrated in Fig.~\ref{fig7} and Tab.~\ref{tab2}, simultaneous prediction is intrinsically challenging, as geometry and appearance exhibit substantially different learning characteristics in terms of optimization difficulty, data distributions, and uncertainty levels. When optimized jointly, these heterogeneous modalities tend to interfere with each other, causing unstable learning dynamics. In practice, this interference not only leads to inconsistent appearance synthesis, such as texture flickering and color instability, but also degrades geometric estimation, resulting in temporally unstable depth predictions with noticeable flickering and structural distortions across frames.

In contrast, our geometry-then-appearance strategy decouples geometric reasoning from appearance synthesis, allowing the model to first converge to a stable and coherent geometric representation. This geometry-aware foundation subsequently provides reliable structural guidance for appearance generation, effectively mitigating cross-modal interference and yielding more consistent depth estimation, more stable geometry, and improved appearance coherence.

\begin{table}[t]
\centering
\small
\setlength{\tabcolsep}{4pt}
\resizebox{\columnwidth}{!}{%
\begin{tabular}{lccccc}
\toprule
\textbf{Method} 
& \textbf{PSNR} $\uparrow$
& \textbf{SSIM} $\uparrow$
& \textbf{LPIPS} $\downarrow$
& \textbf{T-err} $\downarrow$
& \textbf{R-err} $\downarrow$ \\
\midrule
Joint-GTA
& 16.62 & 0.482 & 0.395 & 0.004 & 0.022 \\
\textbf{GTA (ours)}
& \cellcolor{red!25}\textbf{17.47}
& \cellcolor{red!25}\textbf{0.525}
& \cellcolor{red!25}\textbf{0.342}
& \cellcolor{red!25}\textbf{0.001}
& \cellcolor{red!25}\textbf{0.013} \\
\bottomrule
\end{tabular}%
}
\vspace{1mm}
\caption{\textbf{Joint-GTA vs. GTA on the full DL3DV evaluation split.}
GTA outperforms the internal joint-generation variant in both RGB fidelity and pose-based consistency metrics.}
\label{tab:joint_gta_dl3dv_full}
\end{table}

\textcolor{black}{To further provide a more tightly matched internal comparison, we introduce a joint-generation variant termed Joint-GTA. Joint-GTA replaces the serial geometry-then-appearance pipeline with a single diffusion model that jointly predicts RGB and depth, while keeping the rest of the framework as closely matched as possible. Specifically, Joint-GTA uses the same input observations, target camera trajectories, shared frozen video VAE, depth preprocessing, conditioning signals, training data, image resolution, diffusion backbone family, and optimization protocol as GTA. The key difference is that, instead of first generating geometry and then using it as an explicit intermediate condition for a second appearance model, Joint-GTA predicts RGB and depth jointly within a single diffusion process.}

\textcolor{black}{Concretely, we encode the ground-truth RGB video and depth video into latent representations using the same shared VAE, and concatenate the RGB and depth latents along the channel dimension to form a joint latent variable. Since the pretrained video diffusion backbone expects a fixed latent dimensionality, we introduce a lightweight input projection adapter to map the concatenated RGB-depth latent to the backbone input space, while keeping the shared denoising trunk unchanged. At the output side, we use a symmetric lightweight output projection adapter to restore the backbone features to the joint RGB-depth latent dimensionality. The output is then split into RGB and depth branches and passed through two modality-specific prediction heads to estimate the corresponding noise terms. The conditioning branch is kept the same as in GTA, namely the warped partial RGB latents together with the depth latent. In this way, the comparison focuses on stage-wise explicit geometry conditioning versus joint RGB-depth generation while minimizing other implementation differences.}

\begin{table}[t]
\centering
\small
\setlength{\tabcolsep}{4pt}
\resizebox{\columnwidth}{!}{%
\begin{tabular}{lccccc}
\toprule
\textbf{Method} 
& \textbf{PSNR} $\uparrow$
& \textbf{SSIM} $\uparrow$
& \textbf{LPIPS} $\downarrow$
& \textbf{T-err} $\downarrow$
& \textbf{R-err} $\downarrow$ \\
\midrule
Joint-GTA
& 16.34 & 0.463 & 0.418 & 0.005 & 0.023 \\
\textbf{GTA (ours)}
& \cellcolor{red!25}\textbf{17.14}
& \cellcolor{red!25}\textbf{0.502}
& \cellcolor{red!25}\textbf{0.367}
& \cellcolor{red!25}\textbf{0.002}
& \cellcolor{red!25}\textbf{0.016} \\
\bottomrule
\end{tabular}%
}
\vspace{1mm}
\caption{\textbf{Joint-GTA vs. GTA on 20 DL3DV scenes with large viewpoint changes.}
The advantage of GTA remains clear under challenging large-viewpoint settings.}
\label{tab:joint_gta_large_viewpoint}
\end{table}

\textcolor{black}{We evaluate both variants on the official DL3DV evaluation split with 55 scenes under the standard single-pass inference setting without test-time scaling. As shown in Tab.~\ref{tab:joint_gta_dl3dv_full}, GTA consistently outperforms Joint-GTA in both RGB fidelity metrics, including PSNR, SSIM, and LPIPS, and pose-based consistency metrics, including T-err and R-err. Since the two variants share the same backbone family, training data, conditioning inputs, and optimization protocol, this controlled comparison more directly validates the benefit of stage-wise geometry-then-appearance generation.}

\textcolor{black}{We also evaluate the two variants on 20 DL3DV scenes with large camera viewpoint changes, which serve as a more challenging stress-test setting for geometry-aware generation. As shown in Tab.~\ref{tab:joint_gta_large_viewpoint}, the advantage of GTA over Joint-GTA remains clear under large viewpoint changes, where geometry ambiguity and cross-view inconsistency are more pronounced. These results are consistent with our motivation: explicitly generating geometry as an intermediate representation before appearance synthesis provides a stronger structural foundation than jointly predicting RGB and depth within a single denoising process.}

\subsubsection{Random Latent Shuffle.}
We further analyze the effect of Random Latent Shuffle. During training, the input latent features are randomly permuted with probability $p$, which weakens the model’s reliance on view-order correlations inherited from video diffusion pretraining. As a result, the model is less prone to spurious cross-view dependencies and becomes more robust under large viewpoint changes.

\begin{figure}[t]
\centering
\includegraphics[width=0.5\textwidth]{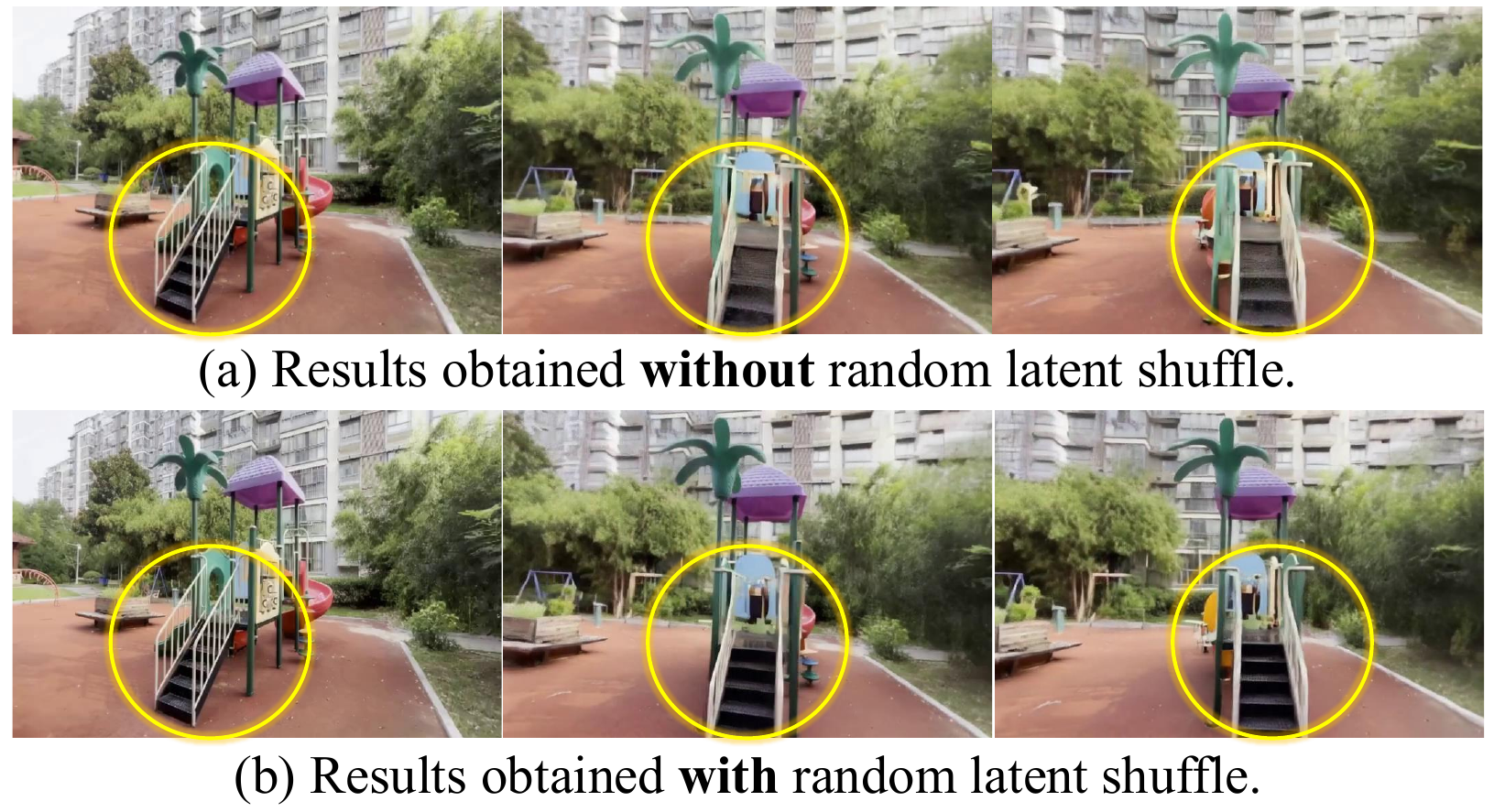}
\caption{\textbf{Qualitative ablation study on random latent shuffle.} Without random latent shuffle, the generated appearance tends to drift across viewpoints, leading to inconsistent textures, as highlighted by the yellow circles. Random latent shuffle mitigates such view-dependent appearance drifting and improves cross-view consistency.}
\label{fig8}
\end{figure}

As shown in Fig.~\ref{fig8} and Tab.~\ref{tab 3}, removing Random Latent Shuffle leads to noticeable appearance drifting across viewpoints. For example, regions that are black in the input image gradually fade into gray in the synthesized views. In contrast, introducing Random Latent Shuffle yields more stable texture statistics and improved cross-view appearance consistency.

\textcolor{black}{This regularization is related in spirit to shuffle-based order-debiasing strategies such as frame shuffling in See3D~\cite{ma2025you}, as both aim to mitigate the order-dependent bias inherited from video diffusion models. However, different from treating shuffling as a default training operation, we activate it with probability $p$, which reduces spurious view-order dependence while preserving the continuity prior from video diffusion pretraining that remains beneficial for coherent generation.}

\begin{table}[t]
\centering
\small
\setlength{\tabcolsep}{4.5pt}
\resizebox{\columnwidth}{!}{%
\begin{tabular}{lccc}
\toprule
\textbf{Training Setting} & \textbf{PSNR} $\uparrow$ & \textbf{SSIM} $\uparrow$ & \textbf{LPIPS} $\downarrow$ \\
\midrule
\textbf{Without} Random Latent Shuffle
& 16.83 & 0.497 & 0.389 \\
\textbf{With} Random Latent Shuffle
& \cellcolor{red!25}\textbf{17.47} & \cellcolor{red!25}\textbf{0.525} & \cellcolor{red!25}\textbf{0.342} \\
\bottomrule
\end{tabular}%
}
\caption{\textbf{Quantitative ablation study on random latent shuffle.} Introducing random latent shuffle during training improves reconstruction fidelity and perceptual quality by alleviating view-order–dependent appearance drifting.}
\label{tab 3}
\end{table}

\textcolor{black}{To further validate this design choice, we conduct a sensitivity analysis on the shuffle probability $p$. As shown in Tab.~\ref{tab:shuffle_prob}, using no shuffle ($p=0.00$) gives the weakest performance. Increasing $p$ consistently improves all three metrics, and the best results are achieved at $p=0.50$. However, applying shuffling to every training sample ($p=1.00$) leads to worse performance than moderate shuffling. This suggests that full shuffling is not optimal in our setting: while shuffling helps reduce order-dependent bias, excessive shuffling may weaken the useful continuity prior inherited from video diffusion pretraining. Therefore, a moderate shuffle probability provides a better balance between strengthening 3D-aware modeling and maintaining coherent cross-view generation.}

\begin{table}[t]
\centering
\small
\setlength{\tabcolsep}{6pt}
\resizebox{\columnwidth}{!}{%
\begin{tabular}{cccc}
\toprule
\textbf{Shuffle probability $p$} 
& \textbf{PSNR} $\uparrow$ 
& \textbf{SSIM} $\uparrow$ 
& \textbf{LPIPS} $\downarrow$ \\
\midrule
0.00 & 16.83 & 0.497 & 0.389 \\
0.25 
& \cellcolor{yellow!25}17.29 
& \cellcolor{yellow!25}0.516 
& \cellcolor{yellow!25}0.356 \\
\textbf{0.50} 
& \cellcolor{red!25}\textbf{17.47} 
& \cellcolor{red!25}\textbf{0.525} 
& \cellcolor{red!25}\textbf{0.342} \\
0.75 
& \cellcolor{orange!25}17.39 
& \cellcolor{orange!25}0.522 
& \cellcolor{orange!25}0.347 \\
1.00 & 17.22 & 0.515 & 0.358 \\
\bottomrule
\end{tabular}%
}
\caption{\textbf{Sensitivity to shuffle probability $p$.}
The best performance is achieved at a moderate shuffle probability ($p=0.50$), whereas full shuffling ($p=1.00$) degrades all three metrics, indicating that excessive shuffling weakens useful continuity priors inherited from video diffusion pretraining.}
\label{tab:shuffle_prob}
\end{table}

\begin{table}[t]
\centering
\small
\setlength{\tabcolsep}{5pt}
\resizebox{\columnwidth}{!}{%
\begin{tabular}{lccc}
\toprule
\textbf{Shuffle strategy} 
& \textbf{PSNR} $\uparrow$ 
& \textbf{SSIM} $\uparrow$ 
& \textbf{LPIPS} $\downarrow$ \\
\midrule
no shuffle 
& 16.83 & 0.497 & 0.389 \\

inconsistent shuffle 
& 16.35 & 0.481 & 0.418 \\

\textbf{consistent shuffle}
& \cellcolor{red!25}\textbf{17.47}
& \cellcolor{red!25}\textbf{0.525}
& \cellcolor{red!25}\textbf{0.342} \\

\bottomrule
\end{tabular}%
}
\caption{\textbf{Ablation study on random latent shuffle strategies.}
Consistent shuffling preserves the correspondence between shuffled latents and warped conditioning signals, while inconsistent shuffling breaks this correspondence and degrades performance.}
\label{tab:shuffle_strategy}
\end{table}

\textcolor{black}{Meanwhile, to verify that Random Latent Shuffle preserves the required correspondences, we compare different shuffling strategies. Specifically, we consider three variants: (1) \textit{\textbf{no shuffle}}; (2) \textit{\textbf{inconsistent shuffle}}, where the noisy latent and the target latent are shuffled with the same permutation while the warped conditioning latent remains in the original order; and (3) \textit{\textbf{consistent shuffle}}, where the noisy latent, the target latent, and the corresponding conditioning signals are all permuted consistently.}

\textcolor{black}{As shown in Tab.~\ref{tab:shuffle_strategy}, inconsistent shuffling performs worse than no shuffle, because although the noisy-target correspondence is preserved, the correspondence between the latent sequence and its conditioning signals is broken. In contrast, consistent shuffling achieves the best performance across all metrics. This confirms that Random Latent Shuffle does not introduce camera/view misalignment in our framework. Instead, when applied consistently to the noisy latent, target latent, and conditioning latent, it preserves all required correspondences while reducing view-order bias, thereby improving cross-view consistency and final synthesis quality.}

\subsubsection{Test-time scaling.}
We further study the effect of the proposed test-time scaling strategy, which aims to improve synthesis quality under large viewpoint changes without introducing additional training cost. 
Although the random latent shuffle strategy significantly enhances cross-view consistency during training, generating high-fidelity novel views at inference time remains challenging when target viewpoints are far from the input view. In such cases, the limited visual evidence provided by the input image yields weak geometric and appearance constraints, often resulting in implausible structures or visually inconsistent results. 

The key intuition behind test-time scaling is that the reliability of synthesized views is not uniform: views closer to the input observation tend to be more accurate, while distant views are more prone to artifacts. Based on this observation, we progressively expand the set of reliable views at inference time and reuse them as additional conditioning signals for subsequent synthesis.

\begin{figure}[t]
\centering
\includegraphics[width=0.5\textwidth]{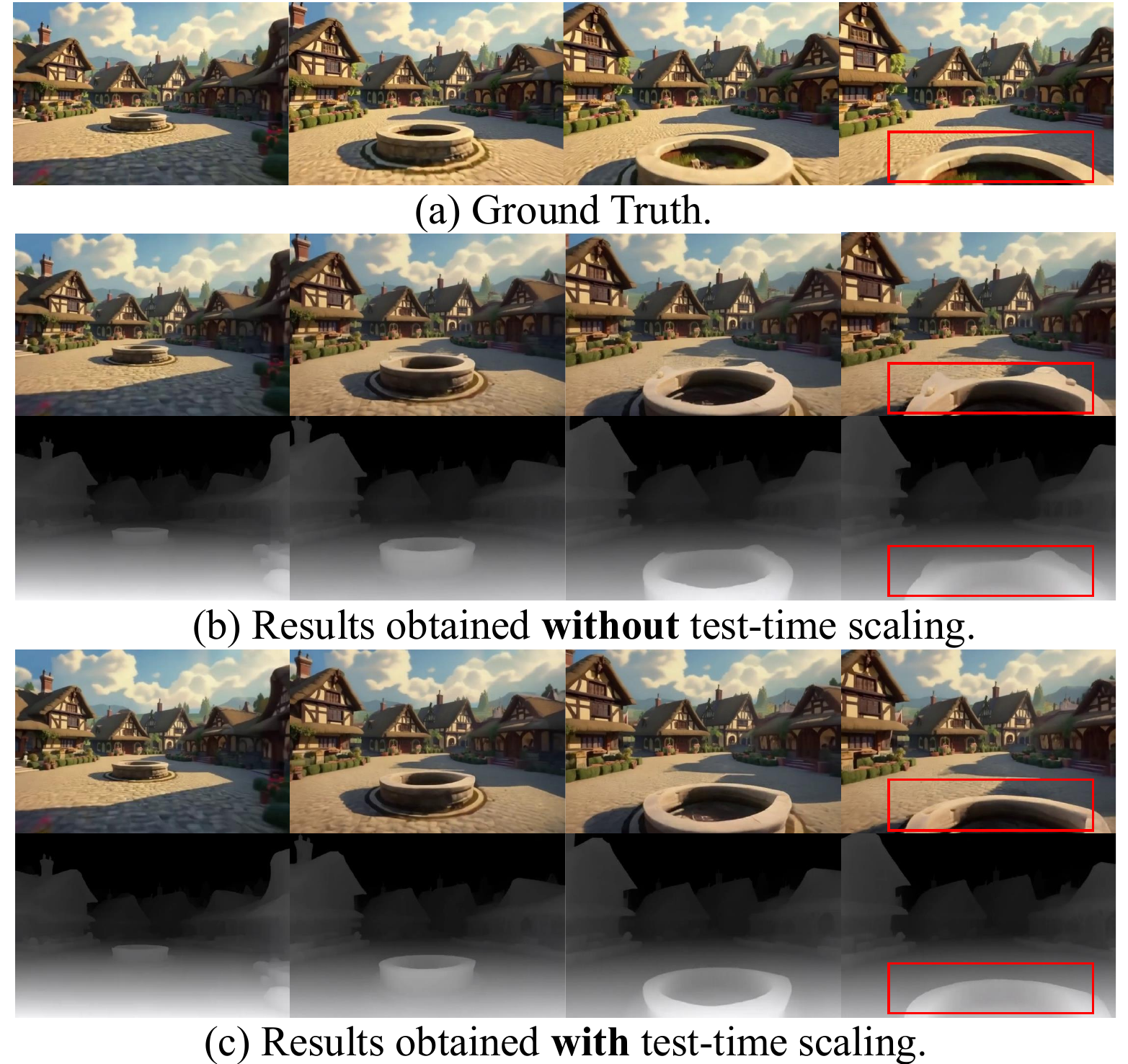}
\caption{\textbf{Qualitative ablation study on test time scaling.} Test-time scaling improves geometric plausibility and visual consistency under large viewpoint changes compared to single-pass inference.}
\label{fig9}
\end{figure}

\begin{table}[t]
\centering
\scriptsize
\setlength{\tabcolsep}{5pt}
\renewcommand{\arraystretch}{1.05}
\resizebox{0.85\columnwidth}{!}{%
\begin{tabular}{lcc}
\toprule
\textbf{Test-time Scaling} & \textbf{Without} & \textbf{With} \\
\midrule
PSNR $\uparrow$ 
& \cellcolor{red!25}\textbf{17.47} & 17.35 \\
SSIM $\uparrow$ 
& 0.525 & \cellcolor{red!25}\textbf{0.531} \\
LPIPS $\downarrow$ 
& 0.342 & \cellcolor{red!25}\textbf{0.340} \\
\midrule
FID $\downarrow$ 
& 38.32 & \cellcolor{red!25}\textbf{34.54} \\
Q-Align-V (quality) $\uparrow$ 
& 3.385 & \cellcolor{red!25}\textbf{3.603} \\
Q-Align-V (aesthetics) $\uparrow$ 
& 2.421 & \cellcolor{red!25}\textbf{2.564} \\
Q-Align-I (quality) $\uparrow$ 
& 3.215 & \cellcolor{red!25}\textbf{3.488} \\
Q-Align-I (aesthetics) $\uparrow$ 
& 2.082 & \cellcolor{red!25}\textbf{2.224} \\
\midrule
T-err $\downarrow$ 
& 0.001 & \cellcolor{red!25}\textbf{0.001} \\
R-err $\downarrow$ 
& 0.013 & \cellcolor{red!25}\textbf{0.011} \\
\bottomrule
\end{tabular}%
}
\caption{\textbf{Quantitative ablation study on test-time scaling.} The proposed strategy consistently improves reconstruction fidelity, perceptual quality, and geometric accuracy.}
\label{tab 4}
\end{table}

\begin{figure*}[t]
\centering
\includegraphics[width=1\textwidth]{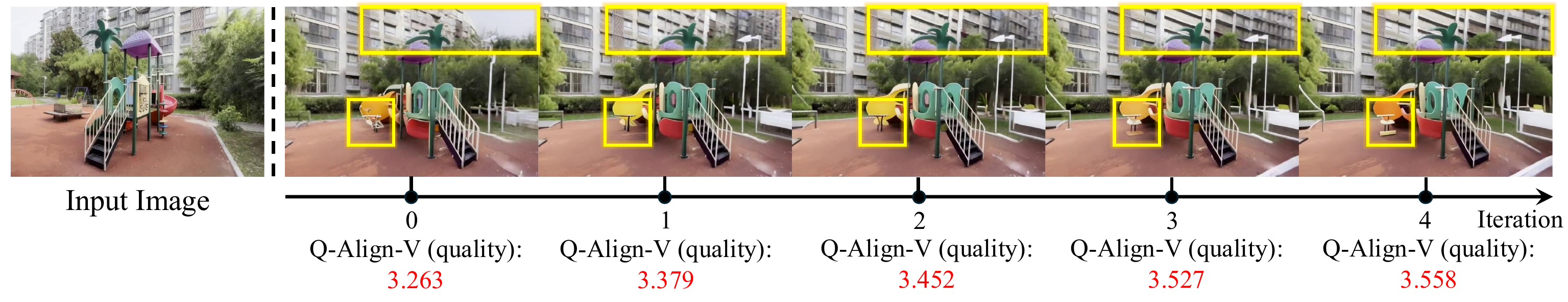}
\caption{\textbf{Visualization of intermediate results of test-time scaling.} As the number of scaling iterations increases, the synthesized results become perceptually more plausible and geometrically more reasonable, while quantitative metrics improve progressively, indicating a stable and effective refinement process.}
\label{fig10}
\end{figure*}

As demonstrated in Fig.~\ref{fig9}, without test-time scaling, the synthesized results often exhibit implausible geometric estimations in regions corresponding to viewpoints far from the input view. In particular, distant views are prone to structural distortions and unstable depth predictions, reflecting insufficient geometric constraints during single-pass inference. In contrast, with test-time scaling enabled, the model produces more geometrically plausible and visually consistent results by progressively incorporating reliable synthesized views as additional conditioning, thereby stabilizing geometry and improving appearance coherence under large viewpoint changes.
As shown in Tab.~\ref{tab 4}, test-time scaling consistently improves performance across multiple metrics, with particularly pronounced gains on all Q-Align scores at both the video and frame levels. These results suggest that the proposed strategy effectively enhances the perceptual plausibility of the synthesized content and achieves closer alignment with human visual perception under large viewpoint changes.

Moreover, as illustrated in Fig.~\ref{fig10}, we further visualize the intermediate results at each iteration of test-time scaling. As the number of scaling iterations increases, the synthesized novel views exhibit progressively improved geometric plausibility and visual consistency, demonstrating that the proposed strategy enables a stable and monotonic refinement process.

\textcolor{black}{We also study the sensitivity of test-time scaling to the reliable-view window size $Q$, which controls how many nearby reliable views are used to guide subsequent refinement. As shown in Tab.~\ref{tab:tts_q_sensitivity}, the best overall performance is achieved at $Q=5$. A smaller window provides insufficient reliable context for challenging target views, while a larger window may introduce less reliable distant views too early and propagate their artifacts to later refinement steps. Therefore, $Q=5$ provides a good balance between reliable conditioning and progressive refinement, and we use it as the default setting in our experiments.}

\begin{table}[t]
\centering
\small
\setlength{\tabcolsep}{4pt}
\renewcommand{\arraystretch}{1.05}
\resizebox{\columnwidth}{!}{%
\begin{tabular}{ccccccc}
\toprule
\textbf{$Q$}
& \textbf{FID} $\downarrow$
& \textbf{Q-V q.} $\uparrow$
& \textbf{Q-V aes.} $\uparrow$
& \textbf{Q-I q.} $\uparrow$
& \textbf{Q-I aes.} $\uparrow$
& \textbf{R-err} $\downarrow$ \\
\midrule
1 & 36.80 & 3.510 & 2.492 & 3.360 & 2.145 & 0.012 \\
3 & 35.10 & 3.570 & 2.536 & 3.450 & 2.198 & 0.011 \\
\textbf{5} 
& \cellcolor{red!25}\textbf{34.54}
& \cellcolor{red!25}\textbf{3.603}
& \cellcolor{red!25}\textbf{2.564}
& \cellcolor{red!25}\textbf{3.488}
& \cellcolor{red!25}\textbf{2.224}
& \cellcolor{red!25}\textbf{0.011} \\
7 & 34.92 & 3.585 & 2.552 & 3.472 & 2.213 & 0.012 \\
9 & 35.38 & 3.561 & 2.538 & 3.446 & 2.197 & 0.012 \\
\bottomrule
\end{tabular}%
}
\caption{\textbf{Sensitivity to reliable-view window size $Q$.}
A moderate reliable-view window ($Q=5$) achieves the best overall perceptual and geometric performance.}
\label{tab:tts_q_sensitivity}
\end{table}

\textcolor{black}{Notably, related iterative inference schemes have also been explored in prior works such as ViewCrafter~\cite{yu2024viewcrafter}, FlexWorld~\cite{chen2025flexworld}, and Gen3C~\cite{ren2025gen3c}. In these methods, iterative generation is primarily used to progressively expand scene coverage or update persistent 3D representations. By contrast, our test-time scaling is designed as an optional refinement strategy for improving synthesis quality under large viewpoint changes, rather than for scene expansion. A key component of this process is the masked-warping mechanism, which distinguishes restored regions, where partial but informative evidence is available, from inpainted regions, where no reliable visual evidence exists. By preserving the former while suppressing unreliable content in the latter before the next refinement round, our method provides cleaner and more stable conditioning for iterative refinement. From this perspective, our method serves as a complement to prior scene-expansion or cache-update strategies, and can be naturally combined with such frameworks. For example, test-time scaling can be used to improve the visual quality of generated results, while prior methods can be used to further extend the spatial coverage of the generated scene}

\textcolor{black}{\subsubsection{Projection module in appearance diffusion.} We further study the projection module used in the appearance diffusion model (Eq.~\ref{eq: appearance video diffusion}) to examine how geometric information should be fused into the appearance branch.}

\textcolor{black}{
Specifically, we compare three variants on the DL3DV dataset:
\begin{enumerate}
    \item \textit{\textbf{w/o residual geometry injection}}, which uses only the projected fused latent, i.e., $\mathbf{h}^{a}=\Pi(\mathbf{z}^{\mathrm{rgb}}\oplus\mathbf{z}^{\mathrm{geo}})$;
    \item \textit{\textbf{cross-attention-based feature fusion}}, which replaces the lightweight projector-based fusion at the same stage with a single cross-attention module. In this variant, the RGB latent $\mathbf{z}^{\mathrm{rgb}}$ is used as the query, while the geometry latent $\mathbf{z}^{\mathrm{geo}}$ is used as the key and value, yielding $\mathbf{h}^{a,\mathrm{CA}}=\mathrm{CrossAttn}(\mathbf{z}^{\mathrm{rgb}},\mathbf{z}^{\mathrm{geo}},\mathbf{z}^{\mathrm{geo}})+\mathbf{z}^{\mathrm{geo}}$;
    \item \textit{\textbf{projector + residual geometry injection}}, our full design, which first aligns the concatenated RGB and geometry latents with the lightweight projection module and then applies residual geometry injection, i.e., $\mathbf{h}^{a}=\Pi(\mathbf{z}^{\mathrm{rgb}}\oplus\mathbf{z}^{\mathrm{geo}})+\mathbf{z}^{\mathrm{geo}}$.
\end{enumerate}
}

\textcolor{black}{As shown in Tab.~\ref{tab:fusion_design}, removing the residual geometry injection consistently degrades performance across all metrics, suggesting that projection alone is insufficient to preserve the structural cues encoded in $\mathbf{z}^{\mathrm{geo}}$. By re-injecting $\mathbf{z}^{\mathrm{geo}}$ after projection, our full design maintains a direct geometry-preserving pathway, which helps the appearance model retain structure-aware guidance under large viewpoint changes. The cross-attention baseline serves as a strong alternative and achieves competitive results, but still performs slightly worse than our full design. We attribute this to the fact that, in our setting, geometry mainly serves as an explicit structural prior for appearance synthesis, and preserving it through a residual pathway offers a simple and stable form of guidance. In comparison, cross-attention introduces a more adaptive but also more indirect interaction between RGB and geometry features. While such flexibility can be beneficial in general multimodal fusion, it may be less effective than explicit geometry preservation for enforcing structure-consistent synthesis in our setting.}

\begin{table}[t]
\centering
\small
\setlength{\tabcolsep}{4.5pt}
\resizebox{\columnwidth}{!}{%
\begin{tabular}{lccc}
\toprule
\textbf{Fusion variant} 
& \textbf{PSNR} $\uparrow$ 
& \textbf{SSIM} $\uparrow$ 
& \textbf{LPIPS} $\downarrow$ \\
\midrule
w/o residual geometry injection 
& 17.29 & 0.519 & 0.349 \\

cross-attention-based fusion 
& 17.43 & 0.523 & 0.344 \\

\textbf{ours}
& \cellcolor{red!25}\textbf{17.47}
& \cellcolor{red!25}\textbf{0.525}
& \cellcolor{red!25}\textbf{0.342} \\

\bottomrule
\end{tabular}%
}
\caption{\textbf{Ablation study of the projection module in appearance diffusion.}
The proposed projector with residual geometry injection achieves the best performance, indicating that explicitly preserving geometric cues benefits structure-consistent appearance synthesis.}
\label{tab:fusion_design}
\end{table}

\textcolor{black}{\subsubsection{Shared VAE for depth encoding.}
We further validate whether the shared frozen VAE can reliably encode depth representations. Since the VAE is pretrained on RGB videos, directly applying it to depth-like inputs may introduce reconstruction distortion. We therefore isolate the VAE encoding-decoding process and evaluate how well depth maps can be reconstructed after passing through the full preprocessing and inverse conversion pipeline.}

\textcolor{black}{Specifically, we obtain depth sequences from DL3DV using Video Depth Anything and denote them as $D$. We then convert the depth maps to normalized disparity, replicate the single-channel disparity maps into three channels, encode and decode them using the frozen video VAE, and finally convert the reconstructed normalized disparity back to depth, denoted as $\hat{D}=\mathrm{Dec}(\mathrm{Enc}(D))$. We compare $\hat{D}$ with the input depth maps $D$ using AbsRel, RMSE, and $\delta_1$, thereby directly measuring the reconstruction distortion introduced by the shared frozen VAE. The metrics are computed over valid pixels following standard definitions: $\mathrm{AbsRel}=\frac{1}{N}\sum_i |\hat{D}_i-D_i|/D_i$, $\mathrm{RMSE}=\sqrt{\frac{1}{N}\sum_i(\hat{D}_i-D_i)^2}$, and $\delta_1=\frac{1}{N}\sum_i \mathbf{1}(\max(\hat{D}_i/D_i,D_i/\hat{D}_i)<1.25)$, where $N$ denotes the number of valid pixels.}

\textcolor{black}{As shown in Tab.~\ref{tab:shared_vae_depth}, the reconstruction distortion is very small, with an AbsRel of $0.016$, an RMSE of $0.019$, and a $\delta_1$ of $0.993$. These results indicate that the pretrained video VAE can encode and decode the normalized disparity representation reliably, supporting its use as a shared tokenizer for both RGB and depth in our framework. This shared latent representation also preserves compatibility with the pretrained diffusion backbone, allowing geometry and appearance features to be modeled within a common latent space.
}

\begin{table}[t]
\centering
\small
\setlength{\tabcolsep}{6pt}
\resizebox{\columnwidth}{!}{%
\begin{tabular}{lccc}
\toprule
\textbf{Setting} 
& \textbf{AbsRel} $\downarrow$ 
& \textbf{RMSE} $\downarrow$ 
& $\boldsymbol{\delta_1}$ $\uparrow$ \\
\midrule
\textbf{shared VAE on depth}
& \cellcolor{red!25}\textbf{0.016}
& \cellcolor{red!25}\textbf{0.019}
& \cellcolor{red!25}\textbf{0.993} \\
\bottomrule
\end{tabular}%
}
\caption{\textbf{Reconstruction error of the shared frozen VAE on depth.}
The pretrained video VAE introduces only minor reconstruction distortion when encoding normalized disparity maps.}
\label{tab:shared_vae_depth}
\end{table}

\textcolor{black}{\subsubsection{Depth Encoding Strategies.}
We further study how depth should be adapted to the shared frozen video VAE. To this end, we compare three depth-to-VAE encodings: raw depth normalization $U_{\mathrm{raw}}$, log-depth normalization $U_{\log}$, and disparity normalization $U_{\mathrm{disp}}$. Let $D$ denote the input depth map, and let $d_{\min}$ and $d_{\max}$ denote the minimum and maximum depth values used for normalization. The three normalized representations are defined as}
$$
\textcolor{black}{
\begin{aligned}
U_{\mathrm{raw}} &= \frac{D - d_{\min}}{d_{\max} - d_{\min}}, \\
U_{\log} &= \frac{\log D - \log d_{\min}}{\log d_{\max} - \log d_{\min}}, \\
U_{\mathrm{disp}} &= \frac{1/D - 1/d_{\max}}{1/d_{\min} - 1/d_{\max}}.
\end{aligned}
}
$$
\textcolor{black}{In all settings, the normalized single-channel representation $U \in [0,1]$ is replicated into three channels and processed by the frozen video VAE using the same preprocessing pipeline as RGB frames. After decoding, each representation is converted back to depth using its corresponding inverse mapping. We then compute the reconstruction error between the input depth and the reconstructed depth. As shown in Tab.~\ref{tab:vae_depth_encoding}, the disparity-normalized representation yields the smallest reconstruction error, indicating that the RGB-trained VAE can reliably tokenize depth-like signals when the depth representation is properly normalized.}

\begin{table}[t]
\centering
\small
\setlength{\tabcolsep}{4.5pt}
\resizebox{\columnwidth}{!}{%
\begin{tabular}{lccc}
\toprule
\textbf{Depth-to-VAE encoding} 
& \textbf{AbsRel} $\downarrow$ 
& \textbf{RMSE} $\downarrow$ 
& $\boldsymbol{\delta_1}$ $\uparrow$ \\
\midrule
raw depth normalized to $[0,1]$ 
& 0.031 & 0.038 & 0.969 \\

log-depth normalized to $[0,1]$ 
& 0.023 & 0.030 & 0.982 \\

\textbf{disparity normalized to $[0,1]$}
& \cellcolor{red!25}\textbf{0.016}
& \cellcolor{red!25}\textbf{0.019}
& \cellcolor{red!25}\textbf{0.993} \\

\bottomrule
\end{tabular}%
}
\caption{\textbf{VAE-level validation of different depth-to-VAE encodings.}
Disparity normalization yields the lowest reconstruction distortion after VAE encoding and decoding.}
\label{tab:vae_depth_encoding}
\end{table}

\textcolor{black}{Subsequently, we evaluate the effect of different depth encodings in the full geometry-then-appearance pipeline on DL3DV. Since changing the depth encoding affects both the target latent distribution of the geometry diffusion model and the geometry-conditioning distribution received by the appearance model, we retrain the corresponding models for each encoding to ensure a fair comparison. For geometry evaluation, we compute AbsRel, RMSE, and $\delta_1$ between the generated video depth and the corresponding reference depth sequence. For final RGB synthesis, we compute PSNR, SSIM, LPIPS, and FID between the synthesized RGB video and the corresponding ground-truth RGB video. As shown in Tab.~\ref{tab:e2e_depth_encoding}, the disparity-normalized representation achieves the best geometry generation quality and also leads to the best final RGB synthesis performance.}

\textcolor{black}{\subsubsection{Training-data controlled comparison.}
To examine the influence of task-specific training data, we conduct a controlled comparison using TrajectoryCrafter~\cite{yu2025trajectorycrafter} as a representative strong baseline. Specifically, we fine-tune TrajectoryCrafter using the same DL3DV training data as GTA, without using RealEstate10K for task-specific fine-tuning. We then evaluate the public checkpoint, the DL3DV-finetuned TrajectoryCrafter, and GTA under the same input/output protocol and metrics on both the DL3DV evaluation split and a held-out RealEstate10K split. This setting controls for the task-specific training data and evaluation protocol.}

\begin{table}[t]
\centering
\small
\setlength{\tabcolsep}{3pt}
\renewcommand{\arraystretch}{1.05}
\resizebox{\columnwidth}{!}{%
\begin{tabular}{lccccccc}
\toprule
\textbf{Depth encoding} 
& \multicolumn{3}{c}{\textbf{Geometry}} 
& \multicolumn{4}{c}{\textbf{RGB synthesis}} \\
\cmidrule(lr){2-4} \cmidrule(lr){5-8}
& \textbf{AbsRel} $\downarrow$ 
& \textbf{RMSE} $\downarrow$ 
& $\boldsymbol{\delta_1}$ $\uparrow$
& \textbf{PSNR} $\uparrow$
& \textbf{SSIM} $\uparrow$
& \textbf{LPIPS} $\downarrow$
& \textbf{FID} $\downarrow$ \\
\midrule
raw depth 
& 0.093 & 0.228 & 0.883 
& 17.25 & 0.513 & 0.354 & 40.4 \\

log-depth 
& 0.089 & 0.221 & 0.892 
& 17.32 & 0.521 & 0.346 & 39.0 \\

\textbf{disparity}
& \cellcolor{red!25}\textbf{0.083}
& \cellcolor{red!25}\textbf{0.212}
& \cellcolor{red!25}\textbf{0.904}
& \cellcolor{red!25}\textbf{17.47}
& \cellcolor{red!25}\textbf{0.525}
& \cellcolor{red!25}\textbf{0.342}
& \cellcolor{red!25}\textbf{38.3} \\
\bottomrule
\end{tabular}%
}
\vspace{1mm}
\caption{\textbf{End-to-end depth encoding ablation on DL3DV.}
Raw depth, log-depth, and disparity are normalized to $[0,1]$. Disparity normalization achieves the best intermediate geometry quality and final RGB synthesis performance.}
\label{tab:e2e_depth_encoding}
\end{table}

\textcolor{black}{As shown in Tab.~\ref{tab:training_data_controlled}, fine-tuning TrajectoryCrafter on DL3DV-only data improves its performance over the public checkpoint on the DL3DV evaluation split, indicating that task-specific training data indeed helps the baseline adapt to this setting. However, under the same DL3DV-only training condition, GTA still achieves substantially better results on both DL3DV and the held-out RealEstate10K split. These results indicate that, under matched task-specific training data, GTA maintains clear advantages on both in-domain and held-out evaluation splits, demonstrating the effectiveness and generalization of the geometry-then-appearance design.
}

\begin{table}[t]
\centering
\small
\setlength{\tabcolsep}{3pt}
\renewcommand{\arraystretch}{1.05}
\resizebox{\columnwidth}{!}{%
\begin{tabular}{lccccc}
\toprule
\textbf{Method} 
& \textbf{Data}
& \textbf{PSNR} $\uparrow$
& \textbf{SSIM} $\uparrow$
& \textbf{LPIPS} $\downarrow$
& \textbf{FID} $\downarrow$ \\
\midrule
\multicolumn{6}{l}{\textit{DL3DV evaluation split}} \\
TrajectoryCrafter (public) 
& public
& 15.82 & 0.454 & 0.412 & 43.94 \\
TrajectoryCrafter (DL3DV-ft)
& DL3DV
& 16.08 & 0.468 & 0.399 & 41.80 \\
\textbf{GTA (ours)}
& \textbf{DL3DV}
& \cellcolor{red!25}\textbf{17.47}
& \cellcolor{red!25}\textbf{0.525}
& \cellcolor{red!25}\textbf{0.342}
& \cellcolor{red!25}\textbf{38.32} \\
\midrule
\multicolumn{6}{l}{\textit{RealEstate10K held-out split}} \\
TrajectoryCrafter (public)
& public
& 16.08 & 0.589 & 0.376 & 32.36 \\
TrajectoryCrafter (DL3DV-ft)
& DL3DV
& 16.02 & 0.592 & 0.371 & 31.90 \\
\textbf{GTA (ours)}
& \textbf{DL3DV}
& \cellcolor{red!25}\textbf{17.01}
& \cellcolor{red!25}\textbf{0.644}
& \cellcolor{red!25}\textbf{0.341}
& \cellcolor{red!25}\textbf{26.72} \\
\bottomrule
\end{tabular}%
}
\vspace{1mm}
\caption{\textbf{Training-data controlled comparison.}
TrajectoryCrafter is fine-tuned with the same DL3DV training data as GTA and evaluated under the same protocol on both DL3DV and held-out RealEstate10K splits.}
\label{tab:training_data_controlled}
\end{table}

\textcolor{black}{\subsubsection{COLMAP-based Pose Evaluation.}
We also evaluate camera-motion consistency using a COLMAP-based SfM protocol on the DL3DV evaluation split. This analysis complements the VGGT-based pose evaluation by providing an alternative pose-estimation protocol based on sparse feature matching and reconstruction. For all methods, we use the same standard COLMAP pipeline, including SIFT feature extraction, feature matching, geometric verification, and sparse reconstruction with default parameters. A generated sequence is considered successfully reconstructed if COLMAP registers at least $80\%$ of its frames and produces a connected camera trajectory. We compute pose errors only on successfully reconstructed sequences and report the COLMAP reconstruction success rate separately.}

\textcolor{black}{Since COLMAP reconstruction is defined up to an arbitrary global coordinate frame and scale, we align the reconstructed trajectory to the reference trajectory before computing pose errors. Specifically, using the frames successfully registered by COLMAP, we estimate a global similarity transformation, including scale, rotation, and translation, that best aligns the reconstructed camera centers to the reference camera centers in the least-squares sense. We then apply this transformation to the reconstructed trajectory and compute T-err and R-err on the registered frames.}

\textcolor{black}{As shown in Tab.~\ref{tab:colmap_pose_eval}, our method achieves the highest COLMAP reconstruction success rate and the lowest pose errors among all compared methods. This SfM-based evaluation is consistent with the VGGT-based results and further indicates that GTA generates videos with stronger geometric and camera-motion consistency.}

\begin{table}[t]
\centering
\small
\setlength{\tabcolsep}{5pt}
\resizebox{\columnwidth}{!}{%
\begin{tabular}{lccc}
\toprule
\textbf{Method} 
& \textbf{COLMAP success} $\uparrow$
& \textbf{T-err} $\downarrow$
& \textbf{R-err} $\downarrow$ \\
\midrule
See3D 
& 84.7\% & 0.006 & 0.025 \\
ViewCrafter 
& 82.9\% & 0.012 & 0.039 \\
TrajectoryCrafter 
& 86.5\% & 0.008 & 0.033 \\
FlexWorld 
& 88.1\% & 0.005 & 0.027 \\
Gen3C 
& 87.3\% & 0.007 & 0.031 \\
Voyager 
& 83.6\% & 0.011 & 0.036 \\
\textbf{Ours}
& \cellcolor{red!25}\textbf{91.4\%}
& \cellcolor{red!25}\textbf{0.003}
& \cellcolor{red!25}\textbf{0.018} \\
\bottomrule
\end{tabular}%
}
\caption{\textbf{Complementary COLMAP-based pose evaluation on DL3DV.}
Our method achieves the highest COLMAP reconstruction success rate and the lowest pose errors, indicating stronger geometric and camera-motion consistency.}
\label{tab:colmap_pose_eval}
\end{table}

\textcolor{black}{\subsection{Geometry Quality Validation}
Since GTA explicitly uses generated geometry as an intermediate condition for appearance synthesis, evaluating the final RGB output alone may not fully reveal the quality and contribution of the geometry stage. We therefore conduct three geometry-oriented evaluations to assess the intermediate geometry from complementary perspectives: direct video-depth estimation, partial-observation geometry completion, and its downstream effect on final RGB synthesis.}

\textcolor{black}{\subsubsection{Direct video-depth evaluation.}
We first conduct a direct geometry evaluation on ScanNet~\cite{dai2017scannet} using 100 RGB-D video sequences constructed from this dataset. \textbf{In this experiment, the input is a full RGB video sequence, and the output is the video depth predicted by our geometry video diffusion model.} We evaluate the predicted video depth against the corresponding ground-truth depth sequence. Following prior video depth estimation protocols, we compute AbsRel, RMSE, and $\delta_1$ between the predicted and ground-truth depth, and compare against representative monocular video depth estimation methods such as Video Depth Anything and DepthCrafter. As shown in Tab.~\ref{tab:direct_geometry_eval}, the intermediate geometry produced by our model remains competitive with recent video depth estimation methods, even though our geometry model is designed as part of an image-to-3D world generation pipeline rather than as a dedicated video depth estimator.}

\begin{table}[t]
\centering
\small
\setlength{\tabcolsep}{6pt}
\resizebox{\columnwidth}{!}{%
\begin{tabular}{lccc}
\toprule
\textbf{Method} 
& \textbf{AbsRel} $\downarrow$ 
& \textbf{RMSE} $\downarrow$ 
& $\boldsymbol{\delta_1}$ $\uparrow$ \\
\midrule
Video Depth Anything~\cite{chen2025video} 
& \cellcolor{red!25}\textbf{0.058} 
& \cellcolor{red!25}\textbf{0.176} 
& \cellcolor{red!25}\textbf{0.949} \\

DepthCrafter~\cite{hu2025depthcrafter}
& 0.064 & 0.186 & 0.940 \\

\textbf{Ours}
& \cellcolor{orange!25}\textbf{0.061}
& \cellcolor{orange!25}\textbf{0.181}
& \cellcolor{orange!25}\textbf{0.944} \\

\bottomrule
\end{tabular}%
}
\caption{\textbf{Direct geometry evaluation on ScanNet.}
Our geometry model achieves competitive depth prediction accuracy compared with dedicated video depth estimation methods.}
\label{tab:direct_geometry_eval}
\end{table}

\textcolor{black}{\subsubsection{Partial-observation geometry completion.}
We further evaluate geometry generation on DL3DV under a setting that is closer to our actual pipeline. Specifically, we select 55 scenes, reconstruct them with 3DGS, and render reference depth sequences for evaluation. \textbf{In this experiment, the input is the partial RGB observation sequence, and the output is the complete video depth generated by our geometry video diffusion model.} We compute AbsRel, RMSE, and $\delta_1$ between the generated video depth and the reference depth sequence rendered from 3DGS, and compare our method with representative baselines for geometry generation from partial observations. As shown in Tab.~\ref{tab:partial_geometry_completion}, our method achieves the best results across all three metrics, with an AbsRel of $\mathbf{0.083}$, an RMSE of $\mathbf{0.212}$, and a $\delta_1$ of $\mathbf{0.904}$. These results directly verify the quality of the intermediate geometry generated by our model in the partial-observation setting required by our pipeline.
}

\begin{table}[t]
\centering
\small
\setlength{\tabcolsep}{6pt}
\resizebox{\columnwidth}{!}{%
\begin{tabular}{lccc}
\toprule
\textbf{Method} 
& \textbf{AbsRel} $\downarrow$ 
& \textbf{RMSE} $\downarrow$ 
& $\boldsymbol{\delta_1}$ $\uparrow$ \\
\midrule
Hunyuan Voyager~\cite{huang2025voyager} 
& 0.101 & 0.238 & 0.869 \\

DualCamCtrl~\cite{zhang2025dualcamctrl}
& 0.109 & 0.252 & 0.852 \\

\textbf{Ours}
& \cellcolor{red!25}\textbf{0.083}
& \cellcolor{red!25}\textbf{0.212}
& \cellcolor{red!25}\textbf{0.904} \\

\bottomrule
\end{tabular}%
}
\caption{\textbf{Partial-observation geometry completion on DL3DV.}
Our method generates more accurate intermediate geometry under the partial-observation setting used in our pipeline.}
\label{tab:partial_geometry_completion}
\end{table}

\textcolor{black}{\subsubsection{Effect of geometry quality on RGB synthesis.}
Finally, we examine whether improved geometry quality translates into better RGB synthesis. To this end, \textbf{we keep the appearance synthesis model fixed and change only the geometry input.} Specifically, we use different methods, including Hunyuan Voyager, DualCamCtrl, and our geometry video diffusion model, to generate video depth from the same partial RGB observations. We then feed each generated video depth sequence, together with the same partial RGB observation sequence, into the same appearance model for RGB synthesis. We evaluate the synthesized RGB videos against the corresponding ground-truth RGB videos using PSNR, SSIM, LPIPS, and FID. As shown in Tab.~\ref{tab:geometry_quality_rgb}, under the same appearance synthesis model, geometry generated by our method leads to the best final RGB synthesis performance across all metrics. In contrast, using geometry produced by Hunyuan Voyager or DualCamCtrl results in lower PSNR and SSIM, higher LPIPS, and higher FID. This controlled comparison shows that the appearance model cannot simply compensate for inaccurate geometry. Instead, higher-quality intermediate geometry directly improves the final RGB synthesis, supporting the effectiveness of our geometry-then-appearance pipeline.}

\begin{table}[t]
\centering
\small
\setlength{\tabcolsep}{4.5pt}
\resizebox{\columnwidth}{!}{%
\begin{tabular}{lcccc}
\toprule
\textbf{Geometry source} 
& \textbf{PSNR} $\uparrow$ 
& \textbf{SSIM} $\uparrow$ 
& \textbf{LPIPS} $\downarrow$ 
& \textbf{FID} $\downarrow$ \\
\midrule
Voyager depth 
& 17.06 & 0.506 & 0.365 & 41.5 \\

DualCamCtrl depth 
& 16.82 & 0.492 & 0.380 & 44.4 \\

\textbf{Ours}
& \cellcolor{red!25}\textbf{17.47}
& \cellcolor{red!25}\textbf{0.525}
& \cellcolor{red!25}\textbf{0.342}
& \cellcolor{red!25}\textbf{38.3} \\

\bottomrule
\end{tabular}%
}
\caption{\textbf{Effect of geometry quality on final RGB synthesis on DL3DV.}
Under the same appearance model, higher-quality geometry from our method leads to better final RGB synthesis performance.}
\label{tab:geometry_quality_rgb}
\end{table}

\textcolor{black}{Overall, these three evaluations jointly show that our geometry stage produces reliable intermediate depth and that such explicit geometry serves as an effective condition for appearance synthesis in the geometry-then-appearance pipeline.}

\textcolor{black}{\subsection{Runtime Analysis}
In this section, we report the inference runtime of different methods under the same hardware setting. All methods are evaluated on a single NVIDIA A800 GPU. For GTA, both the geometry generation model and the appearance generation model use 50 DDIM denoising steps. Generating a 49-frame video at a resolution of $720 \times 480$ takes approximately $2.5$ minutes for the geometry stage and $2.5$ minutes for the appearance stage, resulting in about $5$ minutes per sample in the standard setting.}

\textcolor{black}{Under the same setting, See3D and ViewCrafter take about $2$ minutes per sample, FlexWorld takes about $3.5$ minutes, TrajectoryCrafter takes about $4$ minutes, and Voyager takes about $16$ minutes. These results show that although our two-stage design is more expensive than some single-model baselines, its runtime remains within a practical range and is comparable to several representative diffusion-based methods. Meanwhile, it is substantially faster than heavier alternatives such as Voyager.}

\textcolor{black}{Notably, the runtime reported above does not include test-time scaling. Test-time scaling is an optional inference-time refinement strategy designed for challenging large-viewpoint cases, and is not used in the main quantitative comparison. When enabled, it improves perceptual quality and geometric stability, but introduces additional inference cost that increases approximately with the number of refinement rounds. Therefore, it provides an optional quality--efficiency trade-off rather than being a required component of the standard setting.}

\subsection{\textcolor{black}{Post-hoc} Enhancement}
\begin{figure}[t]
\centering
\includegraphics[width=0.5\textwidth]{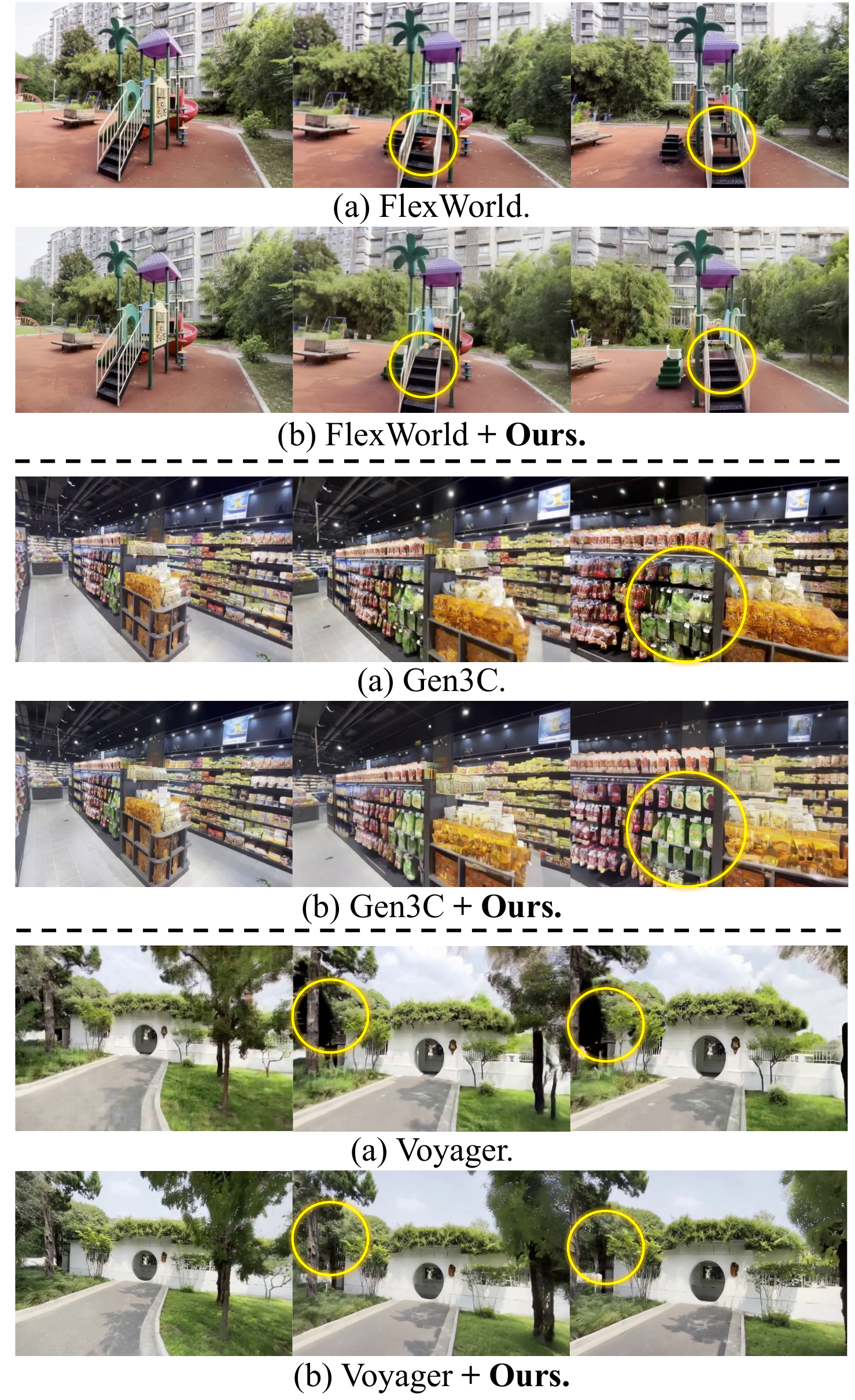}
\caption{\textbf{Qualitative results of post-hoc enhancement.} 
Applying our method as a post-processing module corrects implausible structures and suppresses view-dependent artifacts in the original outputs, resulting in more geometrically plausible and visually coherent generations.}
\label{fig11}
\end{figure}

Beyond serving as a standalone 3D-aware generation framework, our method can also be \textcolor{black}{used as a post-processing enhancement pipeline} for existing image-to-3D scene generation pipelines. Specifically, given cross-view RGB video sequences produced by prior methods, our approach can directly take these outputs as inputs and further refine them through the proposed geometry video diffusion and appearance video diffusion models. \textcolor{black}{In this setting, GTA first performs geometry-oriented processing and then synthesizes appearance conditioned on the refined geometry, thereby introducing an additional geometry-then-appearance inference stage.}

As demonstrated in Tab.~\ref{tab 5} and Fig.~\ref{fig11}, our method consistently improves the quality of results produced by different image-to-3D scene generation pipelines when applied as a \textcolor{black}{post-processing enhancement pipeline}. 
From a quantitative perspective, integrating our approach leads to consistent gains in PSNR and SSIM, along with noticeable reductions in LPIPS, indicating improved reconstruction fidelity and perceptual similarity. From a qualitative perspective, our method effectively corrects implausible structures and suppresses view-dependent artifacts present in the original outputs. 
As shown in Fig.~\ref{fig11}, the enhanced results exhibit more coherent geometry, clearer fine-grained details, and improved cross-view appearance consistency, resulting in visually more plausible and sharper scene generations.

Notably, \textcolor{black}{although the additional geometry generation and appearance synthesis stage incurs extra inference cost,} these improvements are achieved without retraining or modifying the original pipelines, highlighting the flexibility and general applicability of our method.

\begin{table}[t]
\centering
\small
\setlength{\tabcolsep}{4.5pt}
\resizebox{\columnwidth}{!}{%
\begin{tabular}{lccc}
\toprule
\textbf{Method} 
& \textbf{PSNR} $\uparrow$ 
& \textbf{SSIM} $\uparrow$ 
& \textbf{LPIPS} $\downarrow$ \\
\midrule

FlexWorld
& 15.39 & 0.428 & 0.402 \\

FlexWorld \textbf{+ Ours}
& \cellcolor{red!25}\textbf{16.62} 
& \cellcolor{red!25}\textbf{0.488} 
& \cellcolor{red!25}\textbf{0.358} \\

\midrule
Gen3C
& 16.15 & 0.473 & 0.433 \\

Gen3C \textbf{+ Ours}
& \cellcolor{red!25}\textbf{16.83} 
& \cellcolor{red!25}\textbf{0.501} 
& \cellcolor{red!25}\textbf{0.364} \\

\midrule
Voyager
& 15.23 & 0.406 & 0.436 \\

Voyager \textbf{+ Ours}
& \cellcolor{red!25}\textbf{16.98} 
& \cellcolor{red!25}\textbf{0.508} 
& \cellcolor{red!25}\textbf{0.361} \\

\bottomrule
\end{tabular}%
}
\caption{\textbf{Post-hoc enhancement results on different image-to-3D scene generation methods.}
Applying our method as a post-processing module consistently improves reconstruction fidelity and perceptual quality across evaluated baselines.}
\label{tab 5}
\end{table}

\begin{figure}[t]
\centering
\includegraphics[width=0.5\textwidth]{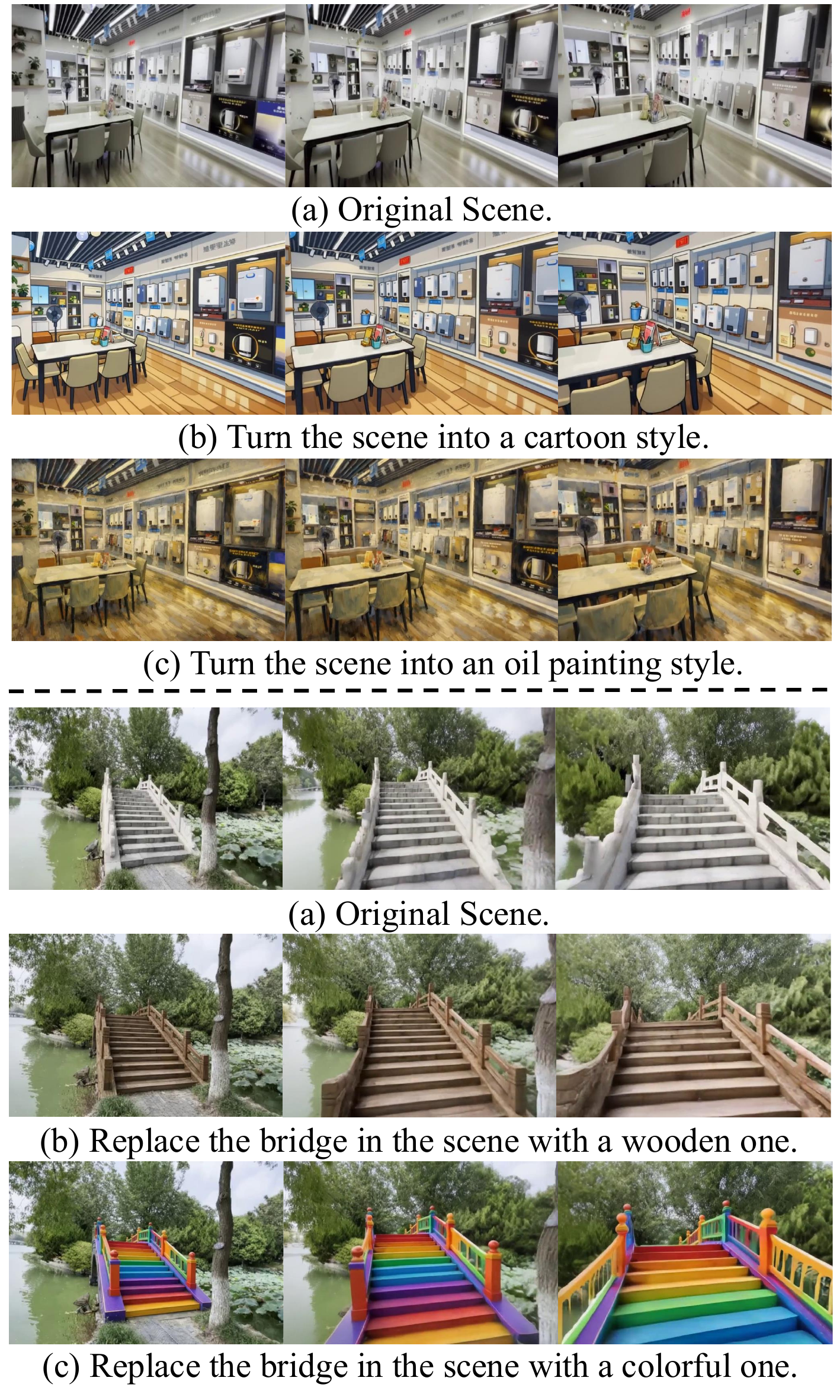}
\caption{\textbf{Qualitative results of 3D scene editing.} 
Our method supports both global appearance stylization (e.g., transforming a scene into cartoon or oil painting styles) and local object-level editing (e.g., replacing a stone bridge with a wooden or colored one), while preserving geometric coherence and cross-view consistency.}
\label{fig12}
\end{figure}

\begin{figure}[t]
\centering
\includegraphics[width=0.5\textwidth]{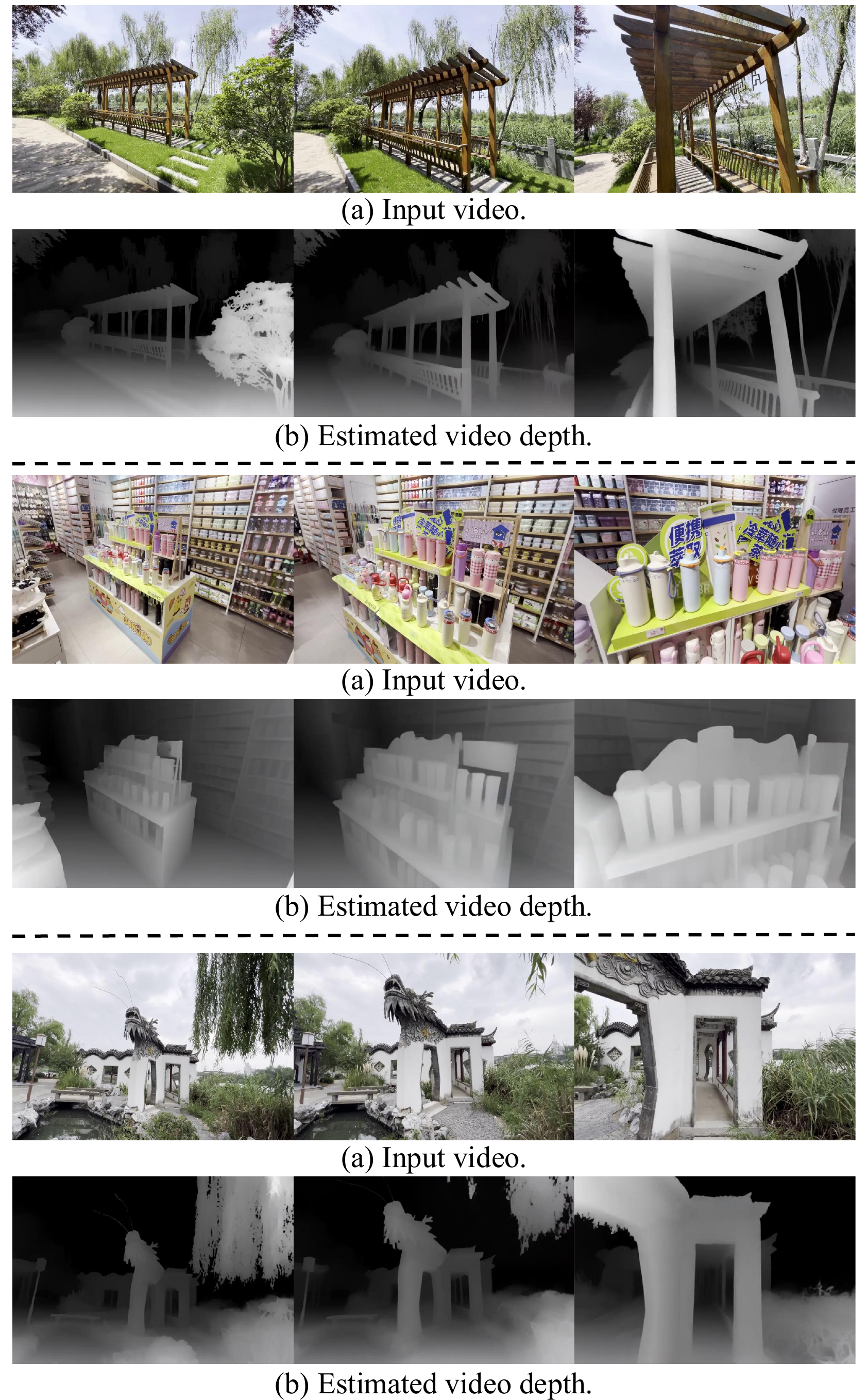}
\caption{\textbf{Qualitative results of video depth estimation.} 
Our geometry video diffusion model produces temporally consistent depth maps that capture fine-grained structural details across frames.}
\label{fig13}
\end{figure}

\begin{figure*}[t]
\centering
\includegraphics[width=1\textwidth]{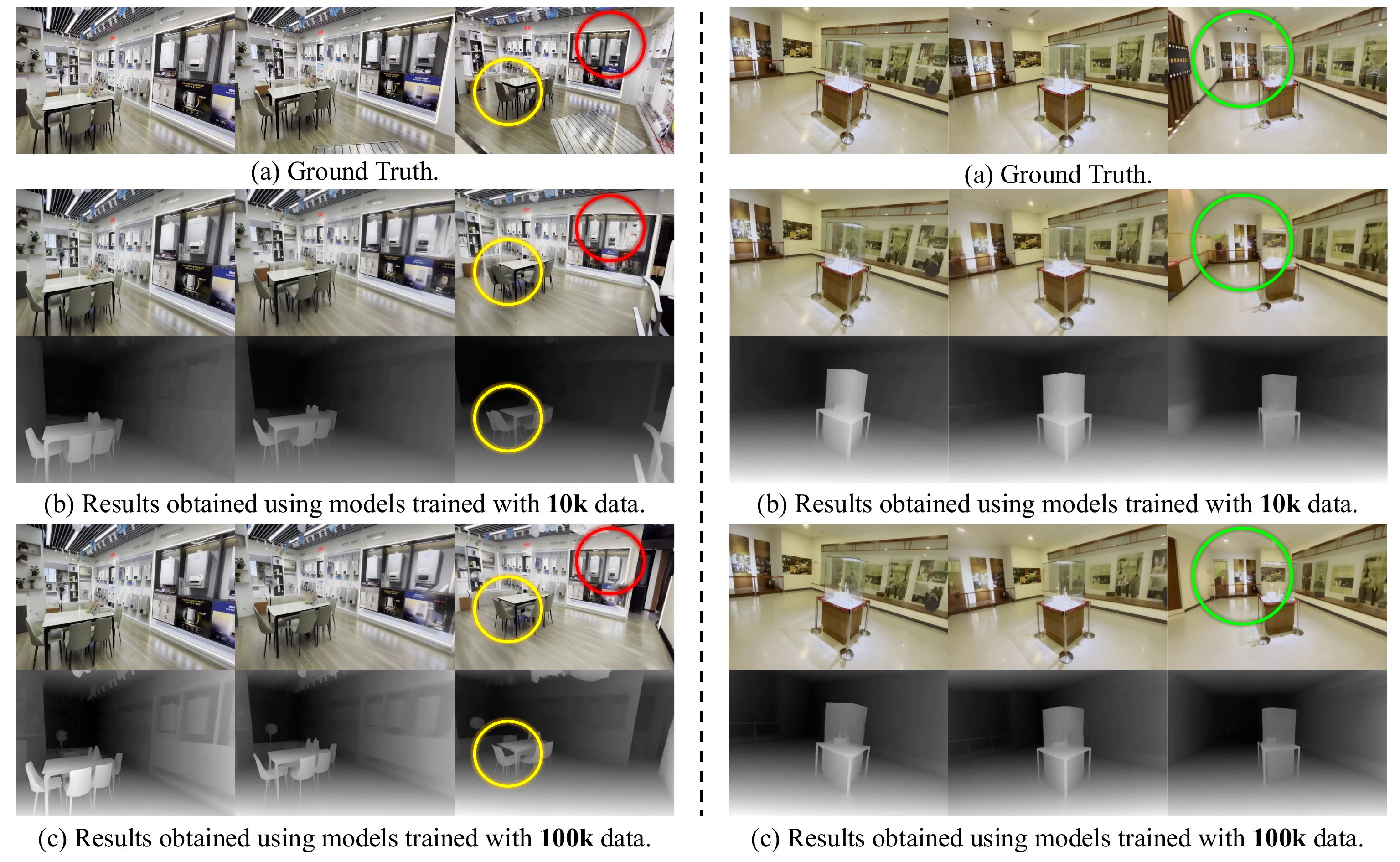}
\caption{\textbf{Qualitative results under different training data scales.} 
Even with only 10K training samples, our method recovers the overall scene structure with reasonable geometric accuracy, while increasing the data scale progressively improves fine-grained details and texture fidelity.}
\label{fig15}
\end{figure*}

\begin{figure}[t]
\centering
\includegraphics[width=0.5\textwidth]{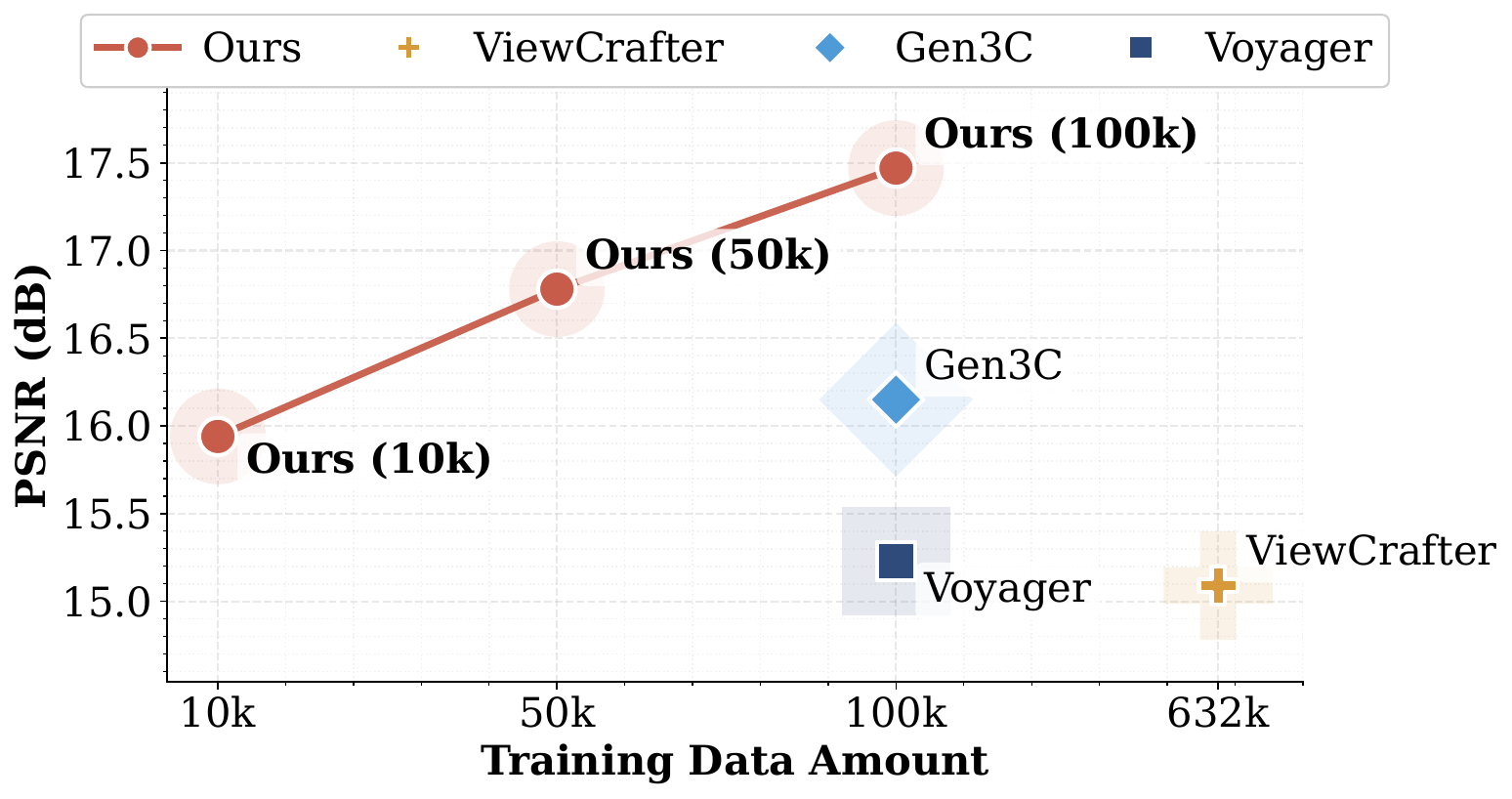}
\caption{\textbf{Data efficiency comparison.} 
Our method achieves comparable or even better performance than state-of-the-art approaches with an order of magnitude fewer training samples, and continues to improve steadily as the training data scale increases.}
\label{fig14}
\end{figure}

\subsection{Applications on Downstream Tasks}
Benefiting from the geometry-then-appearance generation paradigm, our method naturally extends to multiple downstream tasks, such as 3D scene editing and video depth estimation. 
Details are provided below.

\paragraph{3D scene editing.}
To achieve 3D scene editing, we first employ a pretrained image editing model to modify the reference view (i.e., the first frame) of the input scene, while retaining the video depth estimated from the original scene as a representation of the underlying geometry. The edited reference image is subsequently rendered along the original camera trajectories using the preserved depth information to construct a partial RGB sequence. This partial sequence, together with the corresponding depth video, is then provided as conditioning input to our appearance video diffusion model, which synthesizes the edited scene across all target viewpoints. Through this process, the edited appearance is consistently propagated to novel views while preserving geometric coherence.

As demonstrated in Fig.~\ref{fig12}, our approach supports both global and local 3D scene editing in a unified manner. For global editing, the method enables scene-level style transfer, such as transforming an input scene into a cartoon or oil painting style. For local editing, our framework allows targeted modifications of specific scene elements, such as replacing a stone bridge with a wooden or colored bridge, without affecting the surrounding structures. These results illustrate the flexibility of our method in handling diverse editing requirements.

\paragraph{Video depth estimation.}
Our method also naturally supports video depth estimation by leveraging the proposed geometry video diffusion model. 
Given an RGB video as input, the geometry video diffusion model can be directly applied to infer a temporally consistent depth sequence without requiring additional supervision or task-specific training. As illustrated in Fig.~\ref{fig13}, the estimated depth maps exhibit strong temporal consistency across frames and capture fine-grained structural details of the scene, demonstrating the effectiveness of our approach for video-level depth estimation.

\begin{table}[t]
\centering
\small
\setlength{\tabcolsep}{5pt}
\resizebox{\columnwidth}{!}{%
\begin{tabular}{llccc}
\toprule
\textbf{Samples} 
& \textbf{Method}
& \textbf{PSNR} $\uparrow$
& \textbf{SSIM} $\uparrow$
& \textbf{LPIPS} $\downarrow$ \\
\midrule
10K 
& Joint-GTA 
& 15.38 & 0.425 & 0.452 \\
10K 
& \textbf{GTA}
& \cellcolor{red!25}\textbf{15.94}
& \cellcolor{red!25}\textbf{0.463}
& \cellcolor{red!25}\textbf{0.411} \\
\midrule
50K 
& Joint-GTA 
& 16.13 & 0.455 & 0.426 \\
50K 
& \textbf{GTA}
& \cellcolor{red!25}\textbf{16.78}
& \cellcolor{red!25}\textbf{0.501}
& \cellcolor{red!25}\textbf{0.374} \\
\midrule
100K 
& Joint-GTA 
& 16.62 & 0.482 & 0.395 \\
100K 
& \textbf{GTA}
& \cellcolor{red!25}\textbf{17.47}
& \cellcolor{red!25}\textbf{0.525}
& \cellcolor{red!25}\textbf{0.342} \\
\bottomrule
\end{tabular}%
}
\vspace{1mm}
\caption{\textbf{Data-efficiency comparison between Joint-GTA and GTA on DL3DV.}
GTA consistently outperforms the internal joint-generation variant under matched training-data scales.}
\label{tab:data_efficiency_joint_gta}
\end{table}

\subsection{Data Efficiency}

Our proposed method also demonstrates strong data efficiency, enabling reliable geometry generation and appearance synthesis even with limited training data. As illustrated in Fig.~\ref{fig14}, our approach achieves performance comparable to, or even exceeding, state-of-the-art methods while using an order of magnitude fewer training samples. Moreover, as the training data scale increases, our method continues to exhibit steady performance improvements, indicating favorable scalability. As shown in Fig.~\ref{fig15}, even when trained with only 10K samples, our model is able to recover the overall scene structure with reasonable geometric accuracy and visual coherence. As the training data scale increases, the generated results exhibit progressively enhanced fine-grained details and improved texture fidelity, indicating stronger appearance modeling while maintaining stable geometry.

\textcolor{black}{To further examine the source of this data-efficiency advantage, we conduct a controlled comparison between GTA and the internal joint-generation variant Joint-GTA under matched DL3DV training-data scales. Specifically, both variants are trained with 10K, 50K, and 100K samples from the same DL3DV training set and evaluated on the official DL3DV evaluation split with 55 scenes. They share the same frozen video VAE, denoising backbone, conditioning inputs, training data source, and inference protocol. The only difference lies in the generation strategy: Joint-GTA predicts RGB and depth jointly within a single diffusion process, while GTA first generates geometry and then uses the generated geometry as explicit guidance for appearance synthesis.}

\textcolor{black}{As shown in Tab.~\ref{tab:data_efficiency_joint_gta}, GTA consistently outperforms Joint-GTA under all training-data scales. Notably, GTA trained with only 50K samples already surpasses Joint-GTA trained with 100K samples across all three fidelity metrics. This indicates that the proposed geometry-then-appearance strategy uses training data more effectively than directly coupling geometry and appearance prediction in a single diffusion process. Since the backbone, VAE, conditioning inputs, training data source, and evaluation protocol are kept the same, this controlled comparison attributes the data-efficiency advantage more directly to the staged geometry-to-appearance generation design.}

This is consistent with the learning decomposition introduced by the geometry-then-appearance modeling strategy. By prioritizing geometric structure learning, the model first acquires low-frequency, highly regular scene representations that can be reliably learned from limited data, thereby reducing the complexity of the learning task. Subsequently, appearance synthesis is guided by these stable structural priors that reduce ambiguity and narrow the hypothesis space, thus enabling high-quality texture generation with limited data. Together, this formulation leads to a substantially more data-efficient training process.

\textcolor{black}{\subsection{Limitations and Future Work}
Although GTA achieves strong performance in geometry-aware image-to-3D world generation, it still has several practical limitations. First, the current model is mainly trained on DL3DV, which consists primarily of real-world captured scenes. As a result, when the input image falls outside this distribution, such as cartoon-style images, illustrations, or other non-photorealistic content, the generated novel views may exhibit inconsistent visual styles. In particular, the model may preserve the input style for views close to the reference image, but gradually drift toward more realistic textures or inconsistent appearances in distant views. This suggests that the current model is better suited to real-world photographic inputs, while its robustness to stylized or non-photorealistic inputs remains limited.}

\textcolor{black}{Second, the temporal capacity of the underlying video diffusion backbone limits the number of views that can be generated in a single inference pass. In the current implementation, GTA generates a 49-frame multi-view video, which is sufficient for our evaluation protocol and the target-view trajectories considered in our experiments. However, this limits direct application to longer camera trajectories or denser view sampling. Future extensions may incorporate stronger video generation backbones with longer temporal capacity, or adopt autoregressive view generation to support longer-range 3D world synthesis.}

\section{Conclusion}
In this paper, we propose GTA, a novel image-to-3D world generation framework that follows a geometry-then-appearance paradigm. By aligning the generation process with the coarse-to-fine characteristics of human visual perception, GTA adopts a two-stage video diffusion architecture that first establishes structurally coherent 3D geometry and then synthesizes high-quality, view-consistent appearance conditioned on the predicted geometry. To further enhance cross-view consistency, we introduce a random latent shuffle strategy that effectively mitigates view-order–dependent appearance drift during training. In addition, we present a test-time scaling strategy that progressively refines the generated scenes at inference time, improving perceptual quality without introducing additional training overhead. Extensive experiments demonstrate that GTA consistently outperforms existing state-of-the-art methods in terms of structural fidelity, visual realism, and geometric accuracy across multiple benchmarks. Beyond standalone performance, GTA can be naturally incorporated as a post-hoc enhancement module into existing image-to-3D generation pipelines, delivering consistent improvements without architectural modifications. Moreover, the proposed framework generalizes effectively to a range of downstream tasks, including 3D scene editing and video depth estimation. Finally, GTA exhibits strong data efficiency, achieving competitive performance under reduced training data regimes, highlighting its practicality and scalability for real-world 3D generation applications.

\vspace{4mm}
\noindent\textbf{Funding.}
This work was supported in part by the National Natural Science Foundation of China under Grant 62371434 and Grant U25B2010, and by the Zhongguancun Academy under Project C20250302.

\vspace{4mm}
\noindent\textbf{Data Availability Statements.}
The DL3DV dataset~\cite{ling2024dl3dv} is available at \url{https://github.com/DL3DV-10K/Dataset}. The RealEstate10K~\cite{zhou2018stereo} dataset is available at \url{https://google.github.io/realestate10k/}. The code of this paper will be available upon publication.

\noindent\textbf{Conflict of interest.}
The authors have no relevant financial or non-financial interests to disclose.

\bibliography{sn-bibliography}

\end{document}